\documentclass{article}
\usepackage[utf8]{inputenc}
\makeatletter
\renewcommand\paragraph{\@startsection{paragraph}{4}{\z@}%
            {-2.5ex\@plus -1ex \@minus -.25ex}%
            {1.25ex \@plus .25ex}%
            {\normalfont\normalsize\bfseries}}
\makeatother
\usepackage{amsmath}
\usepackage{amssymb}
\usepackage{mathtools}
\usepackage{hyperref}
\usepackage{graphicx}
\usepackage{natbib}
\usepackage{mwe}
\usepackage{widetext}
\usepackage{wrapfig}
\usepackage{lscape}
\usepackage{ragged2e}
\usepackage{float}
\restylefloat{table}
\usepackage{color,soul}
\usepackage[dvipsnames]{xcolor}
\definecolor{blue}{RGB}{0,191,255}
\usepackage{chngcntr}
\counterwithin{figure}{section}
\counterwithin{table}{section}
\usepackage[LGRgreek]{mathastext}
\usepackage{footnote}
\usepackage{enumitem}
\newlist{questions}{enumerate}{2}
\setlist[questions,1]{label=\textbf{RQ\arabic*:},ref=RQ\arabic*}
\setlist[questions,2]{label=(\alph*),ref=\thequestionsi(\alph*)}
\usepackage[toc,page]{appendix}
\usepackage[section]{placeins}
\usepackage{lineno} 

\usepackage{csquotes}
\usepackage{academicons}
\usepackage{tikz,xcolor,hyperref}
\usepackage{authblk}
\usepackage{comment}
\usepackage{tcolorbox}
\usepackage{indentfirst}
\usepackage[font=small,skip=0.5pt]{caption}
\renewcommand{\arraystretch}{1.7} 
\usepackage[hypcap=false]{caption}
\usepackage{subcaption} 
\graphicspath{{./images/}}
\usepackage{booktabs}
\makesavenoteenv{tabular}
\makesavenoteenv{table}
\usepackage{subfiles}
\usepackage{rotating}
\usepackage{adjustbox}

\providecommand{\keywords}[1]
{
  \small	
  \textbf{\textit{Keywords---}} #1
}

\definecolor{lime}{HTML}{A6CE39}
\DeclareRobustCommand{\orcidicon}{
	\begin{tikzpicture}
	\draw[lime, fill=lime] (0,0) 
	circle [radius=0.16] 
	node[white] {{\fontfamily{qag}\selectfont \tiny ID}};
	\draw[white, fill=white] (-0.0625,0.095) 
	circle [radius=0.007];
	\end{tikzpicture}
	\hspace{-2mm}
}

\foreach \x in {A, ..., Z}{%
	\expandafter\xdef\csname orcid\x\endcsname{\noexpand\href{https://orcid.org/\csname orcidauthor\x\endcsname}{\noexpand\orcidicon}}
}

\begin{document}
\pagenumbering{arabic}
\title{Divided by discipline? A systematic literature review on the quantification of online sexism and misogyny using a semi-automated approach}

\author[1]{Aditi Dutta $^{\dagger,\mathsection, *}$ \orcidA{}}
\author[2]{Susan Banducci $^{\dagger}$}
\author[3]{Chico Q. Camargo $^{\dagger,\mathsection,\ddag,\mathparagraph}$}

\affil[1]{ \href{mailto:ad882@exeter.ac.uk}{ad882@exeter.ac.uk}}
\affil[2]{ \href{mailto:S.A.Banducci@exeter.ac.uk}{S.A.Banducci@exeter.ac.uk}}
\affil[3]{ \href{mailto:F.Camargo@exeter.ac.uk}{F.Camargo@exeter.ac.uk}}
\affil[$*$]{  Corresponding author}
\affil[$\dagger$]{\quad University of Exeter, Exeter, United Kingdom}
\affil[$\mathparagraph$]{\quad Oxford Internet Institute, University of Oxford, Oxford, UK}
\affil[$\ddag$]{Ewha Womans University, Seoul, South Korea}
\affil[$\mathsection$]{Alan Turing Institute, London, UK}

\date{} 
\maketitle

\nolinenumbers 

\abstract{Several computational tools have been developed to detect and identify sexism, misogyny, and gender-based hate speech, particularly on online platforms. These tools draw on insights from both social science and computer science. Given the increasing concern over gender-based discrimination in digital spaces, the contested definitions and measurements of sexism, and the rise of interdisciplinary efforts to understand its online manifestations, a systematic literature review is essential for capturing the current state and trajectory of this evolving field.
In this review, we make four key contributions: (1) we synthesize the literature into five core themes—definitions of sexism and misogyny, disciplinary divergences, automated detection methods, associated challenges, and design-based interventions; (2) we adopt an interdisciplinary lens, bridging theoretical and methodological divides across social psychology, computer science, and gender studies; (3) we highlight critical gaps, including the need for intersectional approaches, the under-representation of non-Western languages and perspectives, and the limited focus on proactive design strategies beyond text classification; and (4) we offer a methodological contribution by applying a rigorous semi-automated systematic review process guided by PRISMA, establishing a replicable standard for future work in this domain.
Our findings reveal a clear disciplinary divide in how sexism and misogyny are conceptualized and measured. Through an evidence-based synthesis, we examine how existing studies have attempted to bridge this gap through interdisciplinary collaboration. Drawing on both social science theories and computational modeling practices, we assess the strengths and limitations of current methodologies. Finally, we outline key challenges and future directions for advancing research on the detection and mitigation of online sexism and misogyny.}


\section{Introduction} 

\begin{displayquote}[\citet{secretary2024intensification}]
    \emph{``Millions of women and girls are affected by digital abuse and technology facilitated violence every year. Studies suggest that between 16 and 58 per cent of women have experienced this type of violence.''} 
\end{displayquote}

The rapid growth of online spaces has been accompanied by increased online abuse targeting marginalized groups \citep{wilson2020hate, EURights_2023, vidgen_2020}. Girls and women, in particular, have experienced hostility and harassment in online spaces and platforms \citep{olga_barker_2019, nadim2021silencing, vitak2017identifying}. Research shows that women are twice as likely as men to experience gender-based online harassment \citep{duggan2017online}, often resulting in self-censorship and withdrawal from digital spaces \citep{mantilla2013gendertrolling, EURights_2023, Amnesty_2017}.
This growing concern on the disproportionate impact of online hate speech towards girls and women has given rise to an active interest among the research community in countering online sexism and misogyny \citep{megarry_2014, guest-etal-2021, citron2014hate}, and an increase in research on quantifying the same using machine learning approaches \citep{vidgen_2020}. 

However, these computational approaches face several limitations. They are difficult to implement effectively \citep{hewitt_2016, nozza_volpetti_2019, samory_sen_2021}, as they differ fundamentally in how sexism and misogyny are defined, measured, and operationalized—reflecting the complexity and multiplicity of the underlying concepts \citep{richardson2018woman, matsuda2018public}. 
While these approaches show impressive performance, they fail to identify and capture all forms of sexism or misogyny -- especially overlooking the subtler forms of sexist discourse \citep{rodriguez_2020, rodriguez_2021}, and are often prone to erroneous classifications. This calls for the need to examine the current state of research in online sexism or misogyny, and identifying the current challenges arising due to disciplinary and methodological divide. Compounding this issue is the disciplinary divide: while social science research often adopts qualitative methods to explore sexism in rich, contextualized ways \citep{yasserisexism_2016}, computational work tends to rely on narrow, binary definitions, with limited integration of social science theories to analyze the immense amount of available online data on sexism and misogyny.

Despite growing work, there is no integrative, methodologically rigorous synthesis bridging theoretical frameworks from social science with practical automated computational approaches in computer science.
This gap is particularly critical given that language is a form of social behavior that reflects identity and power structures \citep{dinan_fan_et_al_2020}. Text analysis has been proven to be one of the established methods in mapping and analyzing hostility in online discourses, particularly for online gendered hate-speech \citep{jane2016}. Thus, the need arises to apply a natural language processing (NLP) approach to analyze such data to advance both sociological understanding of the kind of sexism existing in online spaces, and methodological understanding of using and improving computational models to capture the same through detection and identification tasks.

Nonetheless, most of the earlier works have neglected or retrofitted the link between the data and sexism as a theoretical construct \citep{samory_sen_2021, abburi_2021, parikh_2019}. Primarily, sexism and misogyny has been researched as a part of the hate-speech diaspora, disregarding the forms of sexism `not involving hate' \citep{parikh2021}, or other non-hostile forms that are subtle and often deceptive \citep{jha-mamidi-2017, rodriguez_2020, rodriguez_2021, rodriguez_2022}. 
While online communities now emphasize on the detection of sexism [or misogyny] (and other hate speech) more than ever before, automatic detection of these phenomena remains challenging, as most research focuses on using textual features to solve the issue \citep{das_2023}. 

Recognizing this need, \citet{fontanella_2024} conducted a systematic literature review on the study of misogyny using computational methods, where they find a ``limited connection between the areas of knowledge that are necessary to fully grasp this complex phenomenon''. Through our research, we extend the review on articles beyond misogyny with an extensive discussion on the identified practices, along with their challenges and limitations in implementation, backed by social science literature. The review aims to consolidate, categorize, and critique existing approaches to quantifying online sexism and misogyny, identifying key research gaps and challenges through a theoretically informed, methodologically rigorous framework.

To address these gaps systematically, we developed a hybrid review pipeline that combines advanced NLP techniques with traditional PRISMA-based rigor—allowing for scalable yet methodologically transparent analysis of the literature. First, we applied transformer‑based topic modeling (BERTopic) alongside UMAP dimensionality reduction and HDBSCAN clustering to group  abstracts into coherent thematic clusters. Next, we validated and refined these clusters via a KeyBERT‑driven keyword co‑occurrence network, ensuring that selected topics truly reflected our research questions. Finally, we integrated these automated steps with manual title/abstract and full‑text screening to enforce quality and consistency. This hybrid workflow not only accelerates large‑scale literature reviews but also maintains the transparency and reproducibility demanded by systematic review standards.

In this paper, we use \emph{quantification} to refer broadly to the identification, classification, or detection of sexism and misogyny in online texts. Our goal is to map how these constructs are operationalized across disciplines, assess the limitations of existing practices, and pave the way for future research that better integrate theory, method, and application.

To address these gaps, this paper makes four key contributions. First, it synthesizes the fragmented literature across social science and computer science into five central themes: definitions of sexism and misogyny, disciplinary divergences, computational detection methods, current challenges, and design-oriented interventions—addressing the lack of conceptual and methodological cohesion. Second, it bridges disciplinary silos by integrating insights from gender studies, social psychology, and NLP, offering an interdisciplinary perspective on detection models. Third, it highlights critical underexplored areas, such as the absence of intersectional approaches, the dominance of Western contexts and languages, and the limited focus on proactive system design beyond keyword matching. Finally, it provides a methodological contribution through a semi-automated systematic review pipeline, combining topic modeling, network analysis and PRISMA guidelines to establish a transparent and scalable standard for future research in this space.




\section{Background}

\subsubsection*{Defining Sexism and Misogyny: Challenges and Contexts}

Understanding and detecting sexism and misogyny — particularly through automated means — requires careful attention to how these terms are defined. However, definitions of both concepts often prove too narrow to capture their full complexity, especially in computational contexts where nuanced social phenomena must be rendered in operational terms. Despite the central role these concepts play in understanding gendered power relations, there is no cross-disciplinary consensus on their precise definitions. Sexism and misogyny are central concepts in understanding the status of women yet there is no consensus across disciplines on their definition. \citet{wrisley_2023} highlights the difficulty of establishing operational definitions for sexism and misogyny, noting that both terms have evolved well beyond their original conceptual boundaries. 
This ambiguity is compounded in computational fields, where definitional clarity is often sacrificed for operational convenience. Consequently, many studies tailor their definitions to align with their specific research aims, particularly in the detection of hate speech that manifests as sexism, misogyny, or both. 

Manne \citet{manne_2017} provides a foundational framework for distinguishing the two: sexism functions to ``justify [patriarchal] norms, largely via an ideology of supposedly ‘natural’ differences between men and women concerning their strengths  interests, proclivities, and appetites,'' whereas misogyny serves to ``uphold the social norms of patriarchies by policing and patrolling them" indicating it is a systemic property embedded within social structures. While \citet{richardson2018woman} distinguishes between sexist and misogynistic speech, arguing that while sexist speech can oppress without overt violence, misogynistic speech often exhibits key features of hate speech. Building on this systemic view, \citet{srivastava_2017} defines misogyny as a form of hatred or contempt for women, arising directly from patriarchal systems. Traditionally rooted in face-to-face social interactions, misogyny has also functioned historically as a political mechanism to domesticate women, control their sexuality, and undermine collective feminist solidarities \citep{anderson2014modern}.

Some scholars treat sexism and misogyny as synonymous or closely related \citep{rahali_akhloufi_2021, bhattacharya_etal_2020, abburi_2021}, while others conceptualize misogyny as a subset or intensified form of sexism \citep{butt_2021, rodriguez_2020}. Even when the distinction is acknowledged, the terms are frequently used interchangeably in computational studies due to their frequent co-occurrence \citep{frenda_2018}. Much of the computational literature adopts this interchangeable usage, informed by theoretical positions that treat misogyny as an extreme articulation of sexist ideology \citep{chiril_etal_2020, zeinert-etal-2021, kohli_etal_2021}.

Given these definitional challenges, this study aims to synthesize the broad spectrum of existing definitions and examine how sexism and misogyny are conceptualized across both social science and computer research domains.

\subsubsection*{From Offline Harm to Online Hate: The Rise of Digital Sexism and Misogyny}
Back in 2013, the World Health Organization \citep{WHO_2013} identified violence against women as ``a global health problem of epidemic proportions'', primarily referring to offline violence, while also warning of its likely expansion into social media. Indeed, the Internet—particularly social media — has since become a key space for the perpetration of sexism and misogyny, where women are subjected to various forms of violence \citep{olga_barker_2019}. Prior research has also highlighted the role of specific linguistic forms and categories— such as the generic masculine\footnote{A gender-biased form used to indicate those of both masculine and feminine gender, reinforcing a hierarchy that privileges men.} — in reinforcing prejudices, sexist attitudes, and gender stereotypes \citep{sensales_2017}. These manifestations may take different forms but share a common aim: to discredit women’s participation in public life and silence their political voices \citep{olga_barker_2019}. In recent years, systemic gender inequality has increasingly manifested in cyberspace through the proliferation of abusive content that is even more aggressive, prompting further research into this evolving form of online misogyny \citep{fontanella_2024}. As a result, online platforms have contributed to the erosion of boundaries between online and offline experiences \citep{megarry_2014}.

\subsubsection*{Gendered Harassment and the Silencing of Women Online}
Even self-identifying as a woman online can significantly increase the risk of internet harassment. When gender identity is known, gender stereotyping and discrimination from the "real world" often carry over into digital spaces, contributing to a “gender asymmetry” in the dynamics of online abuse \citep{herring_1999}. Even seemingly neutral actions—such as the perceived tone of a post—can be enough to "trigger" misogynistic mockery. Speaking out against such behavior often invites further backlash, with responses that are both sexist and misogynistic in nature, and notably, these can come from both men and women. Those who deliberately derail online feminist spaces often do so to suppress the free speech of those communities \citep{bartow_2009}. \citet{megarry_2014} contextualizes online abuse within discursive practices, arguing that such hostility seeks to silence women’s voices on digital platforms and regulate their public behavior.

The overwhelming volume of gendered abuse online raises serious social concerns. While some victims have been celebrated for exposing abusers through acts of ‘feminist digilantism’ such responses risk reinforcing the notion that these issues should be addressed privately by individuals rather than collectively through systemic or public intervention \citep{jane2016}. Crucially, the impact of misogyny extends beyond psychological harm; it also has material consequences, particularly in how resources and opportunities are distributed in society. Thus, understanding misogyny and gender-based violence in online contexts requires a deeper exploration of their complex entanglement with digital culture and technology—an understanding that is essential to shaping equitable digital gender politics for the future \citep{ging_siapera_2018, ging_siapera_2019}.

\hfill\break
Given its impact, online misogyny and sexism can be seen as ``seeking to prevent women from participating in building the forthcoming technological future''\citep{ging_siapera_2018}. It is therefore necessary to stop such proliferation in online spaces to promote gender equality, raise awareness and eliminate it at the earliest by detecting them through computational tools.

\section{Research questions}\label{sec:res_ques}
The challenge of considering sexism and misogyny from a quantitative perspective, when considering their highly subjective nature, motivates our research questions:
\begin{questions}[leftmargin=.65in]
    \item What are the main topics in the studies identified, and how do they differ by discipline and over time? 
    \item How has the existing literature operationalised sexism and misogyny? 
    \item What are the main challenges and opportunities of computational approaches to the study of sexism and misogyny? Which of the challenges do they address?
\end{questions}
The main objective of this paper is to provide a comprehensive systematic literature review, drawn from the research landscape of sexism and misogyny, studied over the years of 2012-2022. The aim is not to focus on specifics from any individual paper but to provide a general overview of the existing literature and draw conclusions from their study designs and research outputs. These observations are to inspire researchers on best working practices and approaches, while also contributing to future research objectives.

Our systematic literature review is divided into two stages: (a) Identifying the relevant studies through multiple steps by performing a semi-automated selection flowchart as illustrated in the PRISMA flowchart (Figure \ref{fig:prisma}) in Section \ref{sec:identify_rel_study}, (b) Conducting an in-depth analysis of the selected study results in Section \ref{sec:sys_lit_review}. While stage 1 is expected to answer the first research question, stage 2 will answer the second and third research questions.

\section{Identifying relevant studies}\label{sec:identify_rel_study}

\subsection{Search strategy}\label{sec:search_strategy}
\begin{figure}[htbp]
    \centering   
    \includegraphics[width=0.9\textwidth]{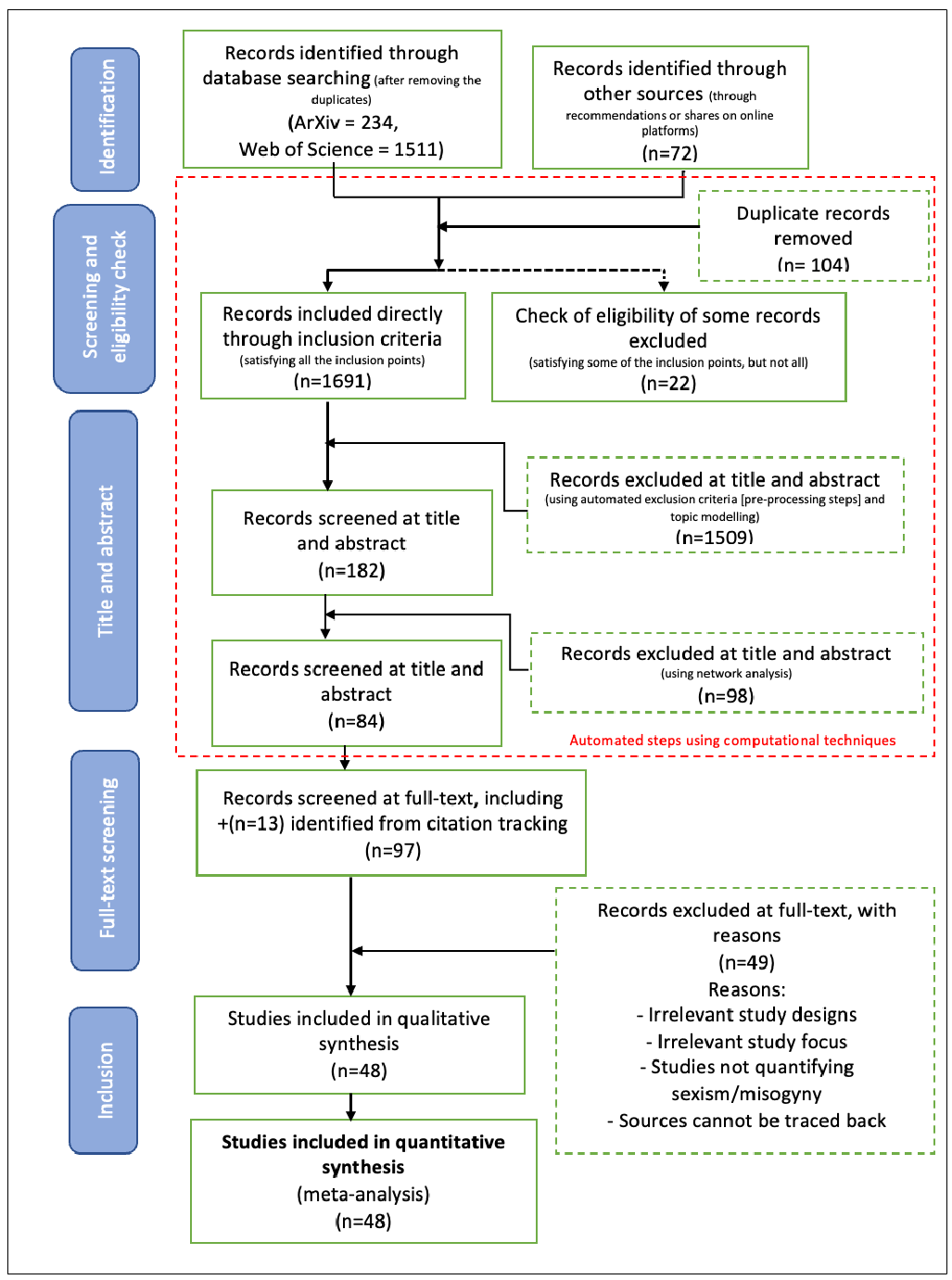}
    \caption{\centering PRISMA flowchart diagram \protect\citep{prisma} for this research. Each step shows the number of studies included and eliminated at that point of the research. }
    \label{fig:prisma}
\end{figure}

We searched six databases -- Google Scholar, ArXiv, Elsevier, Scopus, Semantic Scholar, and Web of Science -- using a closely related set of keywords that operationalized our review criteria of `quantifying' sexism and misogyny. This returned a comfortable number of results that were useful for performing the quantitative analysis. Search results were implemented such that the range of year of publication lay between 2012 and 2022. All of the articles should be in English, containing the full abstracts and titles for each of them. The reporting strategy follows the PRISMA (Preferred Reporting Items for Systematic Reviews and Meta-Analyses), presented in Figure \ref{fig:prisma}\footnote{``The flow diagram depicts the flow of information through the different phases of a systematic review. It maps out the number of records identified, included, and excluded, and the reasons for exclusions. Different templates are available depending on the type of review (new or updated) and sources used to identify studies.''\citep{prisma}}, which uses a checklist approach to systematic literature reviews.

\noindent This research was conducted to review papers with three main characteristics, namely:\label{data_char}
\begin{enumerate}
    \item The papers study sexism and/or misogyny.
    \item The papers ideally study the propagation in social media platforms or other broadcast (preferably text-based) media.
    \item The papers use various methodologies for measuring or quantifying sexism and misogyny (e.g., scales, models, etc.).
\end{enumerate}

ArXiv and Web of Science were chosen to collect studies from the fields of CS and SS, using the search criteria as shown in Table \ref{table:citations}.
As the Figure \ref{fig:prisma} illustrates, the systematic screening and selection process used in this review. An initial total of 1,745 records were retrieved from the databases, comprising 1,511 results from Web of Science and 234 from arXiv. An additional 72 records were identified through external sources during the first stage of the review. Following the removal of 104 duplicates, 1,691 records met the inclusion criteria and were retained, while 22 partially eligible records underwent further assessment and were subsequently excluded. Automated exclusion techniques—comprising text pre-processing, topic modeling, and network analysis—were then applied, resulting in a subset of 84 records selected for title and abstract screening. Subsequently, 97 articles (including 13 from citation tracking) were reviewed at the full-text level. After excluding 49 articles due to irrelevance or lack of methodological clarity, a final set of 48 studies was included for both qualitative and quantitative synthesis. This rigorous process incorporated both manual and computational methods to ensure a focused and comprehensive literature base.

\noindent The data collection method will be discussed in detail in the next section.

\subsection{Data search and collection}\label{sec:data_collection}

In this section, we elaborate on the experimentation conducted with each of the citation databases, and the advantages and disadvantages encountered during the study. For this research, some fields of the search results, namely - title, abstract, year of publication, and the discipline of research for each of the search results were integral to the study. To perform the automated step of narrowing down our search results, some measures were taken to check the consistency and reliability of the data, which is shown in the Table \ref{table:citations}.  

\emph{ArXiv} is a platform that offers researchers to e-publish a draft version of their final work preceding a formal peer review and publication in a peer-reviewed scholarly or scientific journal, also referred to as `pre-prints'\footnote{A preprint is a full draft research paper that is shared publicly before it has been peer-reviewed. Most preprints are given a digital object identifier (DOI) so they can be cited in other research papers. A preprint is a full draft of a research paper that is shared publicly before it has been peer-reviewed. \citep{Mudrak_2018}}. Due to the popularity of ArXiv among CS researchers, its API was used with the expectation of returning unpublished or pre-published works for all disciplines. However, it was found that only the areas of CS vary widely. ``In theoretical computer science and machine learning, over 60$\%$ of published papers are on ArXiv, while other areas are essentially zero." \cite{sutton_2017} We opted to use advanced search queries to narrow down the results, as simpler queries were expected to return more irrelevant results, that had to be removed before analysis. Though the API returned only a  limited number of papers, most of them were found to be relevant. Hence, we took it for analysis but did not use it as our only source due to its skewed disciplinary variety. 

\emph{Web of Science} and \emph{Scopus} showed results in retrieving studies from CS. Though \citet{zhang_2014} found that Scopus retrieved ``significantly'' more studies in CS as compared to the Web of Science, with all of the kinds of document types - conference proceedings, journal articles, reviews, and editorials; yet for our search type, more relevant works were found in Web of Science. As \citet{fiala_tutoky_2017} mentioned in their work, CS has a greater reliance on conference proceedings as compared to other disciplines. To some extent, these conference proceedings papers are also indexed in Web of Science in the Conference Proceedings Citation Index, which makes it possible to carry out scientometric studies of CS based on the data from Web of Science \citep{fiala_tutoky_2017}. 

For \emph{Google Scholar}, we used two external APIs like SerpAPI for scraping the data, as well as a software named \emph{`Publish or Perish'} \citep{publish_or_perish} to collect the search results. Both of the methods were rejected because of their disadvantages. Such as, Publish or Perish could only extract 1000 results at a time for each search query. While this drawback was overcome by searching for documents with a shorter range of years to stay within the limit, it lacked some of the fields that were needed for this study - abstract and discipline. Alternatively, SerpAPI \citep{SerpAPI_2019} worked similar to a web scrapping tool and could only scrape the results as the search engine demonstrates, i.e., it only scrapes what Google shows on their Google Scholar pages, nothing more. Even though the fields we got through this API were relevant, they did not contain the full information we needed for the analysis. For example, the full text in the title and abstract was missing and was instead indicated with dotted extensions in the beginning and end of the text. For the remaining tested citation databases - \emph{Elsevier} and \emph{Semantic Scholar}, the possible search queries were either too simple (consequently giving back a lot of irrelevant studies), did not give back enough studies on our topic, or lacked some of the essential fields (e.g., abstract) that were integral to this study, especially for the automated search strategy used to eliminate non-relevant studies. 

\hfill \break
Therefore, we found empirical evidence indicating that the research outputs we got from ArXiv and Web of Science were ideal for our work. Alongside the search queries, we augmented the dataset with manually added papers that satisfied the selection criteria: \ref{exclusion_inclusion_criteria}. This data from external sources included studies shared in the social platforms Twitter (or X) and LinkedIn, recommendations of other researchers in the field, and following the references of the reviewed papers (i.e., citation tracking).

\subsubsection{Final methodology selection criteria}\label{sec:final_meth_sel_cri}

Observing the pros and cons of all the citation databases, it was decided to use the Web of Science API to collect data based on the individual areas of discipline - SS and CS, as the primary data source. Since many of the relevant computational papers were seen to be published in ArXiv within the given period, those papers were also considered as part of the data collection. It was done to ensure that we get full coverage of both published and unpublished works (pre-prints), relevant to the study of sexism and misogyny during the 11 years. As discussed before, we also included the publications that were informed through external sources. While the Web of Science was taken as the main source for published works, ArXiv was taken as a source for unpublished works. We then combine the selected search results for the next section \ref{sec:data_overview}, before removing the duplicates.

\subsection{Data extraction and synthesis}\label{sec:data_overview}

In this section, we first provide an overview of the collected data from the previous Section \ref{sec:final_meth_sel_cri}, and then use automated approaches for the data extraction stage. The analyses are performed before the application of the selection criteria \ref{exclusion_inclusion_criteria}. For each of the following subsections, the fields considered were:
\begin{itemize}
    \item Title of the paper
    \item Abstract (Multiple abstracts of the same paper were replaced with the first abstract)
    \item Year of publication (or pre-printing)
    \item Language of the paper
\end{itemize}

\begin{figure}[!tbp]
  \centering
  \begin{minipage}[b]{\textwidth} \centering\includegraphics[width=\textwidth,height=6cm,keepaspectratio]{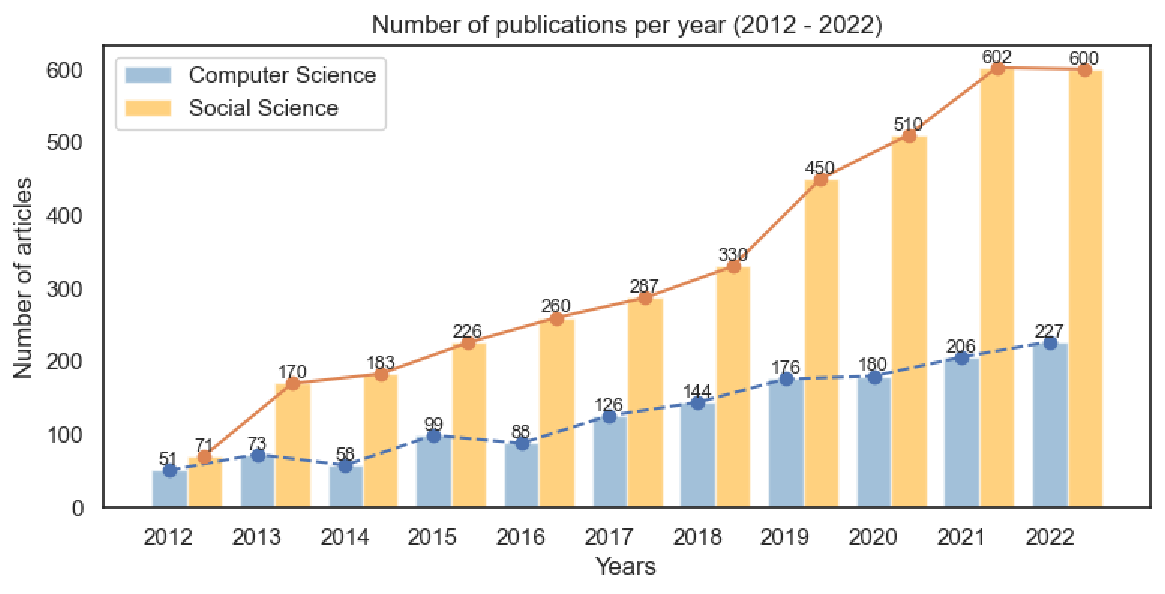}
    \caption{\centering Number of publications per year. \hspace{\textwidth}The blue bars reflect the research articles on Computer Science, while the yellow bars reflect the research articles on Social Science, between the years of 2012-2022.}
    \label{fig:pub_per_year}
  \end{minipage}
  \hfill
  \vspace{-30pt}
\end{figure}

Figure \ref{fig:pub_per_year} shows a steep rise in the study and publication of research on sexism and misogyny, in both the fields of CS and SS. While SS studies always dominated research on the topic, CS works also showed admirable improvement, with a lot of the papers getting published in 2022 alone. The study of online sexism and misogyny has grown significantly since 2014, with a notable rise in scholarly attention from 2018 onwards. Figure \ref{fig:pub_per_year} illustrates the yearly distribution of selected publications, revealing a marked upward trajectory that peaks in 2021.

As we had discussed in the Section \ref{sec:data_collection}, there has been a rising trend of pre-prints in CS \ref{fig:pub_types_disc}, many of which were later published and indexed in citation databases. Studies researching social media platforms like Facebook, Twitter, and Instagram were seen to be limited, with less than 100 works dedicated to research on sexism and misogyny in these online platforms. While almost all of the returned results indicated that works were published majority in English, among the other languages - Spanish and Portuguese followed through, though separated by huge margins.

Pre-processing of the text was done to drop duplicates and remove characters in the text that could hinder the automated selection of the studies based on the titles and abstracts. Studies containing no abstracts at this stage were removed as they could not be added for automated selection criteria. Given that the count of such papers was only 13, the abstracts were looked up in Google Scholar and later manually checked, if they satisfied the selection criteria for this research. 

For the automated extraction stage, we perform two steps in chronological order: \textbf{topic modeling} and \textbf{keyword co-occurrence network} to narrow down our search.

\subsubsection{Topic modeling}\label{subsec:topic_modeling}

Topic modeling\footnote{``Topic modeling is a machine learning technique that automatically analyzes text data to determine cluster words for a set of documents. It leverages `unsupervised’ machine learning to analyze and identify clusters or groups of similar words within a body of text'' \citep{Pykes_2023}.} was used with the pre-processed data containing the abstracts and titles from both disciplines, to generate clusters of topics based on the documents (i.e., the collection of studies containing abstracts and topics). Among all the topic modeling techniques experimented with, BERTopic \citep{grootendorst_2022} proved to be the best choice for the task. It is because BERTopic ``leverages transformers and c-TF-IDF to create dense clusters allowing for easily interpretable topics whilst keeping important words in the topic descriptions''\citep{grootendorst_2022}, hence enhancing the topic recognition ability by the model.\footnote{More information on the techniques used in this methodology is explained in the supplementary section: \ref{sec:topic_mod_approach}}

\begin{figure}[!tbp]
  \centering
  \begin{minipage}[b]{\textwidth} 
  \centering\includegraphics[width=0.9\textwidth,keepaspectratio]{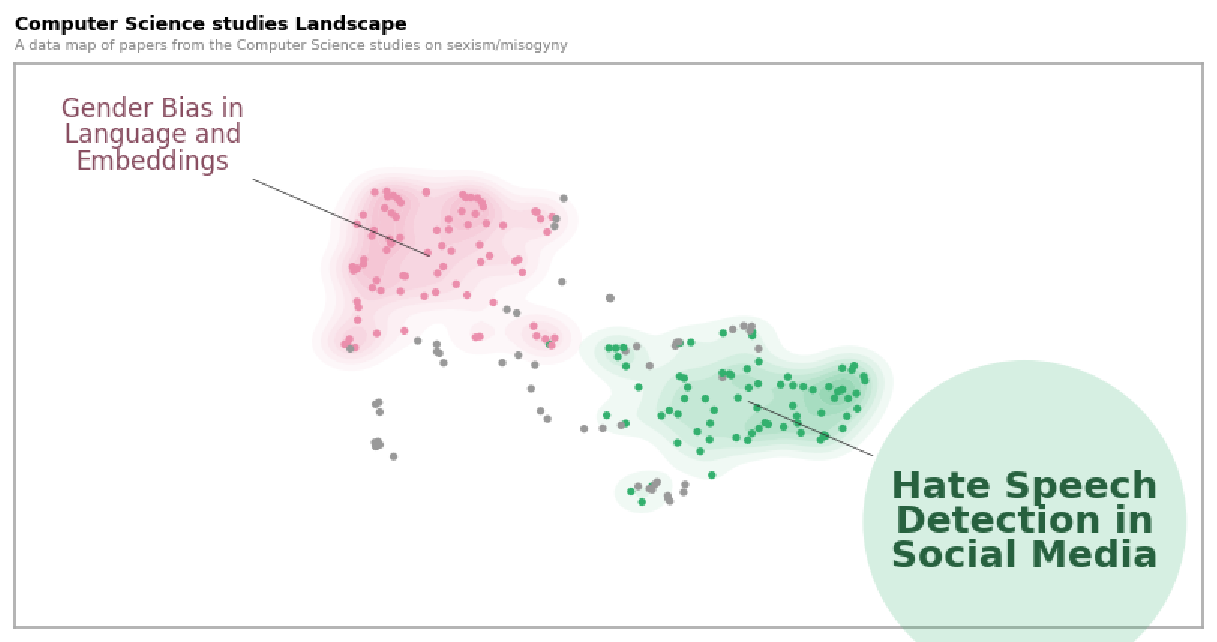}
  \end{minipage}
  \caption{\centering This figure show a UMAP scatterplot, where each point represent one \hspace{\textwidth}document. The unique colors in the figure represent a different  topic in computer science \hspace{\textwidth}centering around sexism and misogyny between 2012 and 2022. Through topic modeling, \hspace{\textwidth}usually each document get assigned a set of key words as themes within the paper, which are then grouped together with an unique color, representing the same topic with similar sets of keywords found across all the documents. When grouped, each topic is described by their topic name in the same color. The grey points represent outliers (documents which did not get any \hspace{\textwidth}assigned topic). The highlighted topic name indicates more relevance to our research objectives.}
  \label{fig:comp_sci_topics}
\end{figure}

\begin{figure}[ht!]
  \centering
  \begin{minipage}[b]{\textwidth} \centering\includegraphics[width=\textwidth,keepaspectratio]{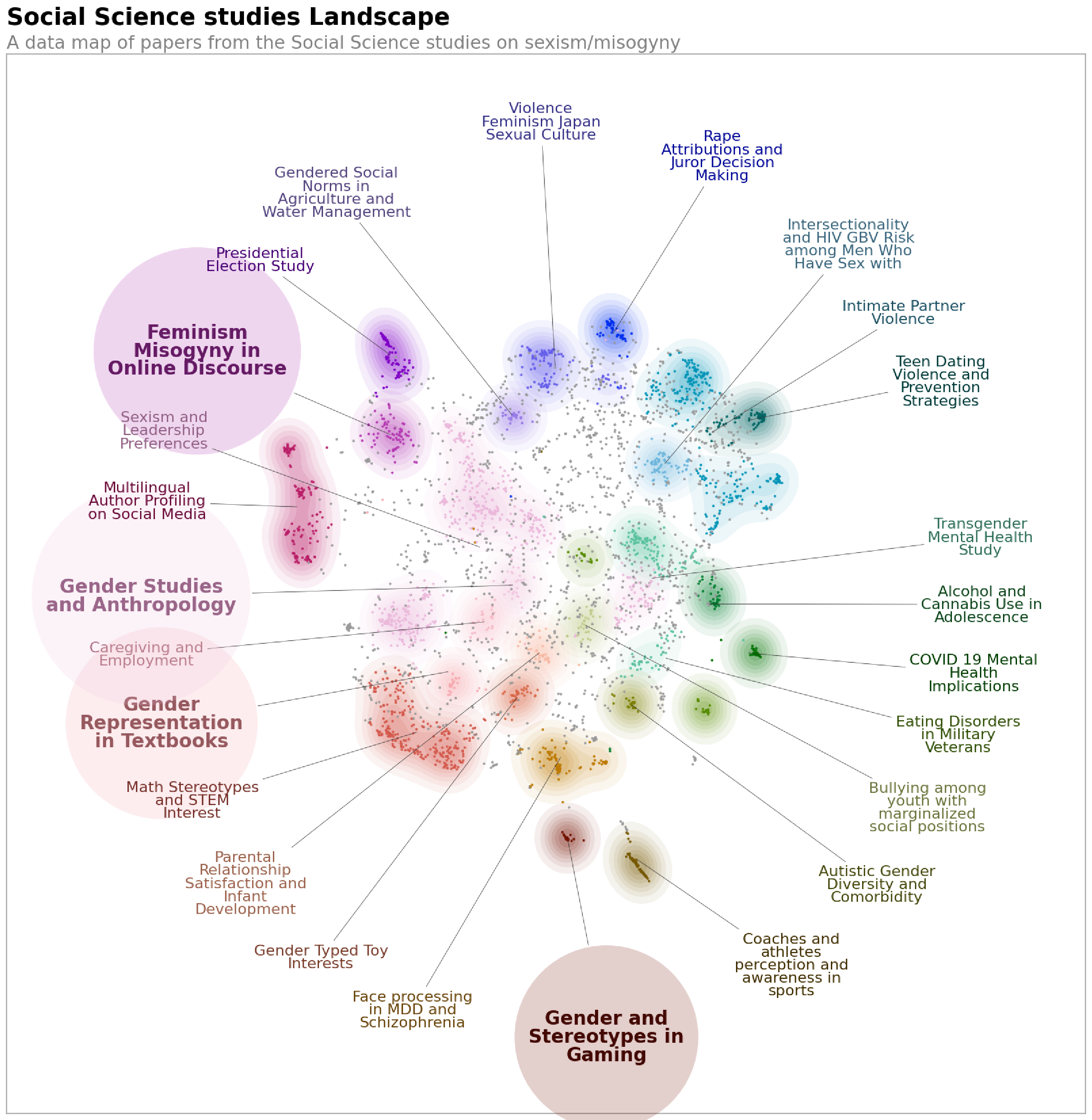}
  \end{minipage}
    \caption{\Centering Similar to Figure \protect\ref{fig:comp_sci_topics}, this figure show a UMAP scatterplot where each each unique color represent a different topic in social science centering around sexism and misogyny \hspace{\textwidth}between 2012 and 2022. The highlighted topic indicates more relevance to our research objectives.}
    \label{fig:soc_sci_topics}
\end{figure}

We applied the BERTopic algorithm to the collections of CS and SS papers separately to capture the topics of research between these two disciplines, and to check the differences in the themes of sexism between them. Among all the experimentation conducted - including setting different ranges of parameters to get the best representative models from there, we further employed fine-tuning of the model to improve on that, by using multiple representations from the model. For our work, we used these different representations from keywords and phrases to summarize and custom labels. The Figures \ref{fig:comp_sci_topics} and \ref{fig:soc_sci_topics} indicate the topics recognized by the model. Using the aforementioned parameters, we used the BERTopic model to groups documents into topic clusters, identified by their keywords and keyphrases. It uses clustering to define topics and hence does not assign more than one topic to each document. In the figures, each point corresponds to each document in their respective disciplines. BERTopic uses HDBSCAN by default for clustering, and it does not force all the data points to be a part of any of the recognized clusters or topics. Simultaneously, BERTopic uses UMAP to perform dimensionality reduction. We then used further customization of the UMAP by setting the parameter `n\_components' to 2, to `pre-reduce' embeddings for visually depicting our model results in the two figures. For those topics that do not form a part of any groups (also termed as ``outliers''), the points are marked in grey in the figures. The colored points in both the figures indicate topics, and each color represents a unique topic for the sets of documents, which have further been marked correspondingly with labels of the same color boxes. Larger clusters represent more densely populated or well-defined research areas, indicating a richer or more mature body of literature. The algorithm itself exhibits strong local clustering to group similar topic categories together, to which we also controlled the balance between the local and the final structure to efficiently distinguish between each topic. 

It uses a light hue of the same colors encircling each topic to indicate the cluster belonging to the respective topic.
In the figures  \ref{fig:soc_sci_topics} and \ref{fig:comp_sci_topics}, different colors indicate different cluster of topics, as identified by the model. Usually in topic modeling, the models analyze ``bags'' or groups of words together to capture the meaning of the words. However, in our approach, we used the BERTopic framework using a Large Language model (LLM) -- Mistral for better inference and contextual understanding of the abstracts and generation of logical topics (indicated by each color).
While some points in the same cluster may look further away than the points from another cluster, it is due to its projection in 2D-dimensional space which we did for better visualization; hence the points within the same clusters are closer in a multi-dimensional space. 

The highlighted circles represent the most relevant thematic areas for our study. Figure \ref{fig:soc_sci_topics} illustrates key domains within social science research, including digital feminism and online misogyny, gender representation in education and media, health-related concerns such as mental health and gender-based violence, and legal-cultural perspectives on sexual violence. The left side of the map emphasizes discourse and identity, while the right side is more focused on public health and psychological issues. The central regions reflect interdisciplinary intersections, suggesting potential for integrating discourse analysis with clinical and educational research to deepen understanding of gendered bias.
Figure \ref{fig:comp_sci_topics} illustrates the computer science research landscape on sexism and misogyny, revealing two dominant and distinct thematic clusters. The first, ``Gender Bias in Language and Embeddings'', centers on studies that investigate algorithmic bias in NLP models, focusing on how word embeddings and language representations encode and perpetuate gender stereotypes. The second cluster, ``Hate Speech Detection in Social Media'', includes work on building and evaluating systems for identifying misogynistic and abusive language online. Unlike the diverse and interdisciplinary structure observed in social science literature, the computer science landscape appears more siloed, with limited overlap between technical fairness research and applied detection systems. This highlights a relatively narrow scope and methodological uniformity in the field's engagement with gendered bias.
We observe that social science studies generally lack research focused on the automated detection or identification of online sexism and misogyny -- the area that this study aims to address. Therefore, we chose to further investigate the articles in Computer Science.

Online sexism and misogyny are subsets of online hate speech. Therefore, in this study, we focus on the topic within computer science that specifically addresses the quantification of such content, rather than broader analyses of gender bias in its various forms (which was one of the two topics identified through topic modeling). To quantify hate speech, it is essential to review research that emphasizes detection tasks. Therefore, at this stage, we selected the topic \textbf{`Hate Speech Detection in Social Media'} from among the various categories and disciplines considered. In the next step, we further demonstrate the relevance of this chosen topic by conducting a keyword co-occurrence network analysis.

\subsubsection{Keyword co-occurrence network}\label{subsec:keywrd_co}

To validate if the topics captured from the automated selection of topics from each discipline in the previous section were representative of the corresponding documents, we navigated the disciplines and each topic, alongside their respective keywords. To obtain the most frequent keywords in the set of documents, we use KeyBERT\footnote{KeyBERT is a minimal and easy-to-use keyword extraction technique that leverages BERT embeddings to create keywords and key phrases that are most similar to a document.}\citep{grootendorst2020keybert} to extract embeddings with a BERT model to get a document-level representation from our abstracts and titles. From each document, we used KeyBERT to identify key phrases that would provide with a more accurate summary of the documents, rather than simple keywords. KeyBERT works by creating an embedding of document texts, from which BERT key phrase embeddings of a pre-defined word n-gram range length of 1-2 words\footnote{Word n-gram range lets users decide the length of sequence of consecutive words that should be extracted from a given text.} were created. Consequently, cosine similarities between the document and their respective keyphrase embeddings are calculated to extract the top 10 keyphrases that best describe that document. These selected keyphrases per document are then compared against the whole set of documents. We chose to look into the 100 most common keywords in the documents taken at both discipline level (CS and SS), as well as the topic level (each topic based on the topics we generated in Section \ref{subsec:topic_modeling}). This was done to check the relevance of the keywords, and consequently the set of documents that would best represent our research objective of performing a literature review on the quantification of online sexism and misogyny.

\begin{figure}[ht!]
    \centering   
    \includegraphics[width=0.8\textwidth, keepaspectratio]{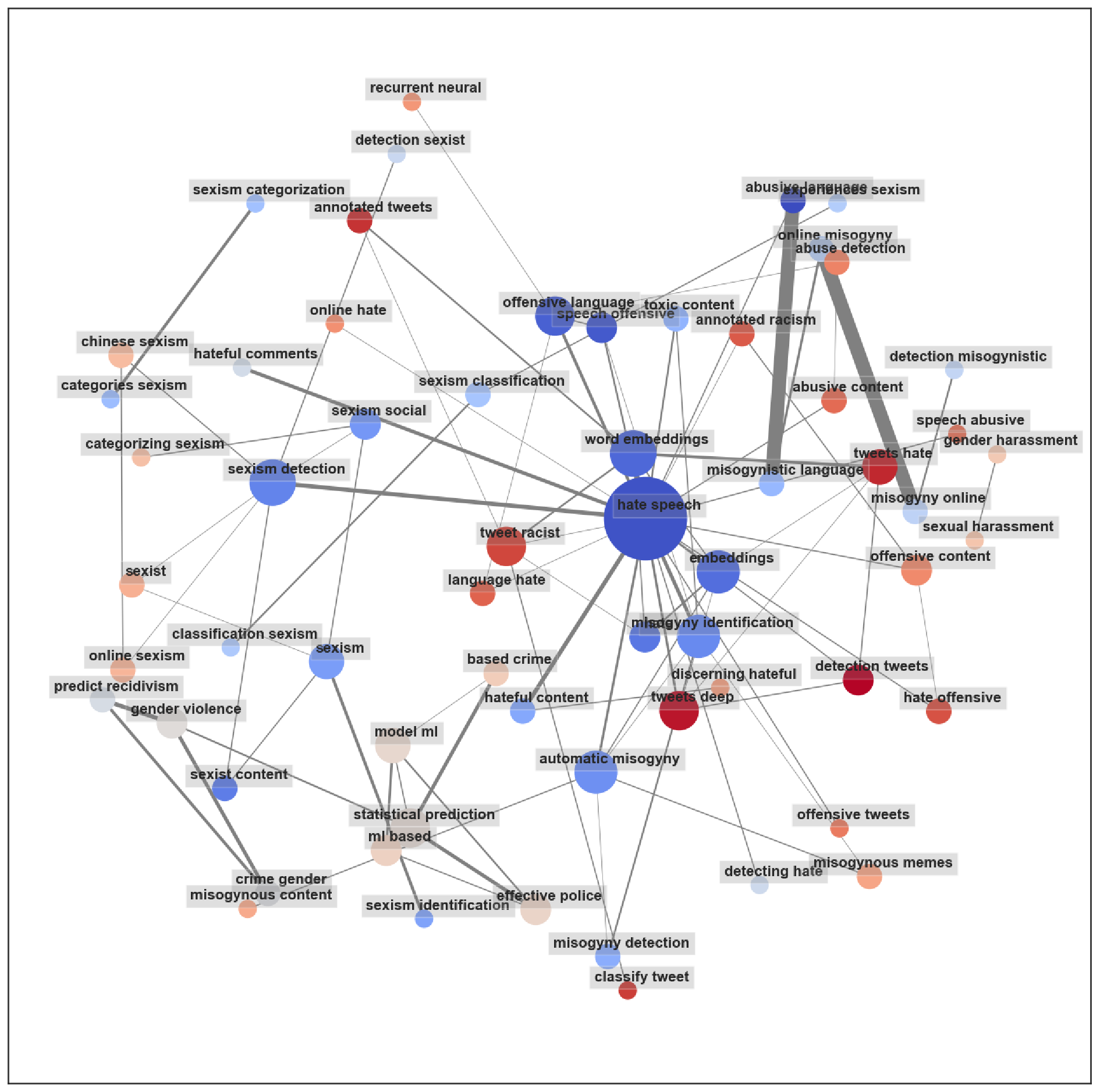}
    \caption{\centering Most frequent keywords gathered from the abstracts and titles of Computer Science\hspace{\textwidth} studies in the topic of \textbf{`Hate Speech Detection using Deep Learning models'}}
    \label{fig:keywords_sel_comp}
     \vspace{-5pt}
\end{figure}

On comparing the keywords present in Computer Science and Social Science, we found that the contents of the papers (from their title and abstract) focused on different kinds of sexism and misogyny - both indicating similarity in topics, but contribution at different capacities. They reveal a complementary yet fragmented landscape in how sexism and misogyny are studied across disciplines.  In Computer Science (see Figure \ref{fig:keywords_comp}), research tends to cluster tightly around specific psychological constructs—such as gender stereotypes, hostile and benevolent sexism, ambivalent sexism, general sexist attitudes, and the broader term ``sexism'' itself—reflecting a strong focus on operationalizing and detecting discrete forms of bias in online text. In contrast, Social Science scholarship (see Figure \ref{fig:keywords_soc}) embraces a broader, more intersectional view: central themes include gender stereotypes and roles, but extend into areas such as gender equality, identity (including transgender experiences), domestic and sexual violence, critical discourse, and systemic discrimination. 

This divergence suggests that computational approaches risk oversimplifying the complexity of sexism by concentrating on narrowly defined categories, while social-theoretic work may lack the granular tools needed for large‑scale detection. Bridging these perspectives would offer a path forward: embedding rich, contextualized constructs from social science themes of power, identity, and structural inequality into algorithmic models to enhance their sensitivity to subtle, intersectional manifestations of sexism and misogyny. Conversely, employing automated keyword and topic‑modeling techniques to capture emergent patterns (e.g., discourse around political discussions) can help social scientists process vast corpora and refine theoretical frameworks. Ultimately, a synergistic integration—where computational precision is being guided by theoretic depth from social science will yield more robust, fair, and actionable tools for understanding and combating online sexism. We will discuss more on this in Section \ref{sec:challenges}.

On further analysis, we selected the most relevant topic among all of the highlighted topics from both disciplines and performed a keyword search on each of them. 
Figure \ref{fig:keywords_sel_comp} corresponds to the topic `Hate Speech Detection in Social Media', and it showed the most promising result of containing the necessary keywords needed for this study. 

The figure consists of the 100 most frequent keyphrases in the topic. The size of each circle (node) indicates the weight of that particular keyphrase in the set of documents. The color of the nodes can be anything that is recognized by Matplotlib in colormap specified and could be randomly generated. The edge width of the edges joining the nodes indicates the number of associations between the two connected nodes - the more the number of connections, the thicker the edge width. The co-occurrence network for all the highlighted topics from the previous step showed us the most used keywords pertaining to the said topics. By studying the generated keywords, we could capture the general objectives of the respective papers under the same topic, as stated from their abstracts. We then selected the most relevant topic for our work, which focuses on computational approaches in detecting online sexism and misogyny (a part of the online hate speech discourse). Consequently, we decided to perform the full-text literature review on the articles which fell under the topic of `Hate Speech Detection in Social Media'.

On completing this step, we included a few more of the studies that had been identified through citation tracking and did a full-text screening of all the collected texts. \\

\noindent\textit{Note: As we see from the plots, till here, all of the analyses largely relied on the information provided in the title or the abstract of the papers. Hence, it limits us in providing a concise assessment of the exact models, methodologies, or datasets used by the corresponding papers. For which we would need a full-text assessment.}

\subsection{Final data selection}\label{sec:final_sel}

\subsubsection{Screening}\label{sec:screening}
Following the selection of the citation database, the automatic filtering of papers using BERTopic, and the validation via topic keywords, we identified the necessary characteristics \ref{data_char} of the data to emphasize the findings which eventually led us to select the research articles based on the selection criteria: \ref{exclusion_inclusion_criteria}, by performing a manual screening. Eligible articles were divided into two categories at this stage. The first category is the data acquired from the automated stage, while the second category is the records identified from citation tracking.

Finally, we thoroughly reviewed all publications related to the quantification of sexism and misogyny in online social platforms to determine their focus and methodologies, by reading their full text.

\subsubsection{Qualitative assessment of the selected studies}\label{subsec:analysis_sel_studies}
A total of 96 full-text articles were analyzed qualitatively, as shown in Figure \ref{fig:prisma}. We assessed them based on \emph{four} criteria, namely: 
\begin{enumerate}[itemsep=0mm,label=(\roman*)]
    \item \emph{Irrelevant study focus} - Whether they are focused on studying the propagation of sexism and misogyny (irrespective of whether they indicated such in the abstract). Many of the works focused on hate speech, but because we wanted to only review sexism and misogyny, they were eliminated.
    \item \emph{Irrelevant study designs} - Whether the intended outcome of the research was not about performing a computation analysis on the detection or identification of online sexism and misogyny.
    \item \emph{Studies not quantifying sexism/misogyny} - Whether the paper focused on a review of studies in the relevant topic; or contained a summary of the author's thoughts from multiple papers, such as opinion pieces around the same topic. We only focused on methodology papers which described their approach. Hence, summary of methods, shared task descriptions or dataset papers\footnote{Our dataset analysis was based on the same set of papers selected during the screening process.} are generally excluded from the list. 
    \item \emph{Sources could not be traced back} - in case the paper's paraphrased contents with citations were not reflective of the summary the original authors indicated in their study.  
\end{enumerate}
45 out of the 96 research articles qualified from this step, from surpassing these exclusion criteria and as a result, were included in the final qualitative analysis.

The full text of all the papers was reviewed qualitatively and information about each was added to a summary table covering the following points:
\begin{itemize}[itemsep=0mm]
    \item Forms of hate speech studied. It is because hate speech could encompass a lot of things, including sexism and misogyny.
    \item Definitions of sexism or misogyny (or both) used in the study.
    \item Language(s) of the data used for the study.
    \item Data selection criteria. This could depend on the original data collection method, such as - using keywords, hashtags, public profiles by monitoring user's online activity, users identified as sexist/misogynist, tags of sexism, specific phrases; or even based on a particular timeline of interest.
    \item Datasets used and their types (external, API-generated, etc.)
    \item Dataset modifications, if done. This could be in the form of data augmentation, counterfactual examples, document expansion by adding semantically similar words, transliterating multilingual dataset to uniform language, and many more.
    \item Broadcast media or social media platform which is of interest for the study.
    \item Annotators used in the study, and their tasks. If each or group of annotators had different tasks, that was also recorded.
    \item Pertaining to the previous point, the Kappa values that are statistical measures used to measure inter-rater reliability, are also noted.
    \item Research bias addressed or acknowledged in the study. If acknowledged, it is posted as a limitation in the paper.
    \item Pre-processing or post-processing done on the data.
    \item Performance metric used.
    \item Embedding type used, since this could range from word-level to node-based.
    \item Classification or clustering type, and the respective models.
    \item Syntactic, linguistic, and semantic/lexical features. 
    \item Prompt topics and intersectionality (if present).
\end{itemize}

\hfill\break
Recently several SLR tools have incorporated semi-automation using Artificial Intelligence techniques, for supporting the screening and extraction (pre-screening) phases \citep{bolanos_2024}, like we did in our research. Of such tools, only a few use topic modeling for their work. Such as, RobotAnalyst\footnote{\url{https://www.nactem.ac.uk/robotanalyst/}} and SWIFT-Review\footnote{\url{https://www.sciome.comswift-review/}} uses Latent Dirichlet Allocation (LDA) that assigns a topic to a paper based on the most recurrent terms shared by other papers using a generative probabilistic model, while Iris.ai\footnote{\url{https://iris.ai/}} clusters the papers according to a two-level taxonomy of global topics and specific topics \citep{bolanos_2024}. The former two tools depend on the term frequencies while the later perform Named Entity Recognition (NER) and allow users to to customize entity extraction by letting them define their own set of categories beforehand. Even with its advantage of the superior language capabilities to produce one of the most advanced techniques in language topic modeling today \citep{Briggs_2023}, BERTopic has remained unexplored for the same task. In our work, we use that potential alongside the promising result of Large Language Models (LLMs) in their information extraction capabilities, to cluster our topics before validating the results with network analysis and selecting the topic(s) more suited for our work. This proved to be particularly useful to us in the screening and qualitative assessment phase as empirical analysis of the topics generated to their corresponding papers showed that the approach accurately clustered similar papers together.

\section{Results of the Systematic Literature Review}\label{sec:sys_lit_review}

\subsection{Data statistics}\label{sec:data_stats}

Post data screening and qualitative assessment, we finally narrow down the number of manuscripts to 45, that satisfied the scope of our meta-analysis. In the first subsection, we provide a brief overview of the key statistics of the 45 papers. In the following subsections, we provide an overview of the existing computational approaches dedicated to quantifying sexism and misogyny. Beyond that, we discuss the challenges and limitations faced by the said approaches from the existing literature.

\subsubsection{Author collaboration network}
\begin{figure}[htbp]
    \centering   
    \includegraphics[width=0.9\textwidth, keepaspectratio]{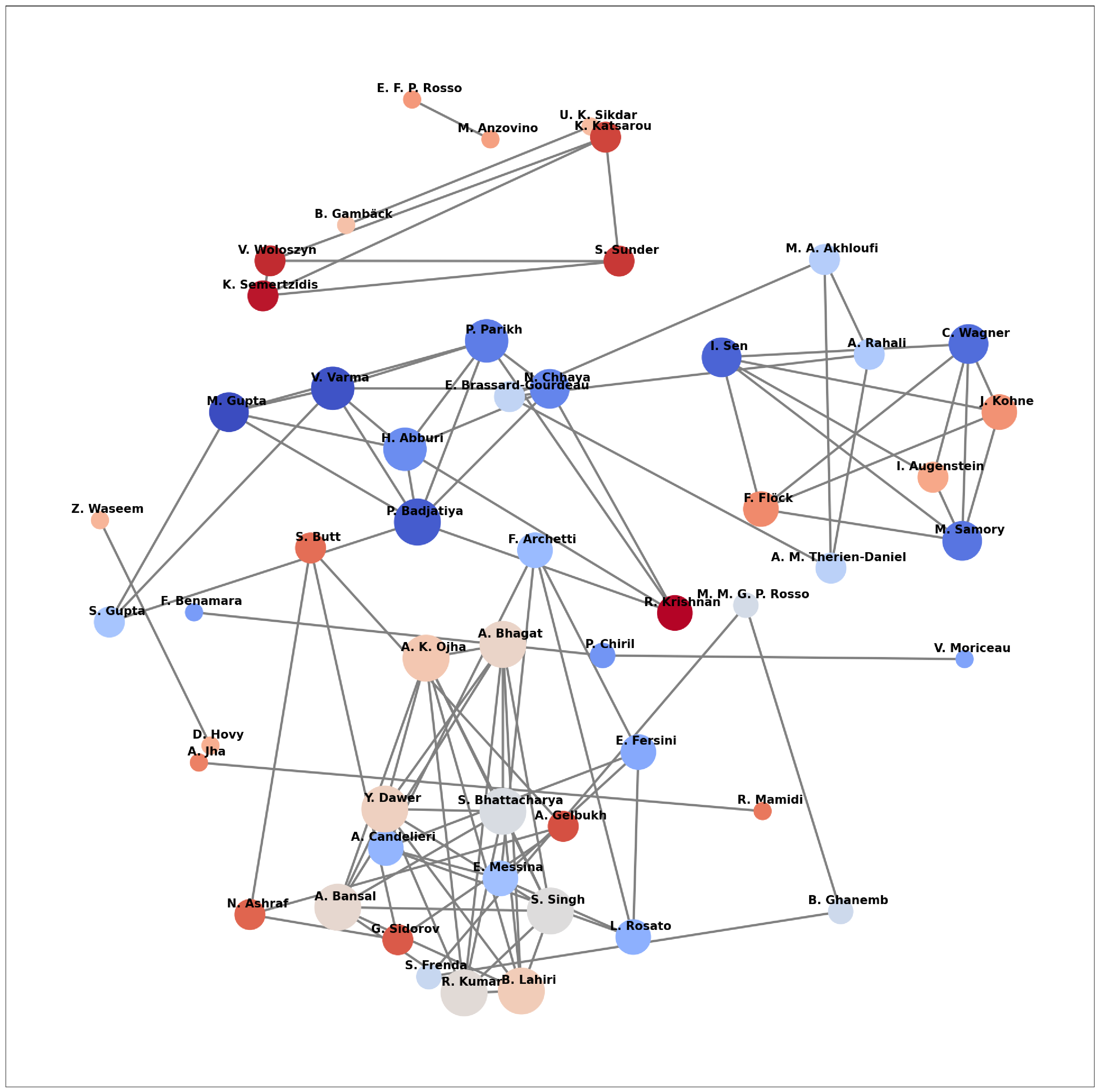}
    \caption{\centering Network displaying all the author collaborations between the 45\hspace{\textwidth} research articles screened for our literature review.}
    \label{fig:auth_network}
\end{figure}

We provide an author collaboration network in Figure \ref{fig:auth_network}, where the name of the researchers are nodes, their size and color indicating the number of relevant manuscripts they authored or co-authored. The connections between the authors are indicative of co-authorship on manuscripts, and their weighted edges imply the frequency of co-authorship.

\subsubsection{Characteristics of study or research designs}
Table \ref{table:sys_obs} gives a summary of all the general characteristics we found in the full-text reviews of the selected studies. It provides a summary of the most used categories of each field (design/methodology) that the documents in our literature review have used. The other categories which were not featured in the table were mostly used by only one document. Each document could have multiple categories under the same field of design or methodology. For example, one document could be researching datasets of multiple languages in different platforms of interest and using multiple models, with different levels of classification at different stages. The categories that are uniquely present in a document are marked with an asterisk($^\ast$) beside it, while the fields with their entire list of categories in the table are marked with an obelisk($^\dag$). In here, the asterisk($^\ast$) symbol would not just indicate that the feature itself is unique to the field, but also all the documents should add up to the total number of literature listed.

\begin{table}[ht!]
    \centering
    \renewcommand{\arraystretch}{1.1}
    \begin{tabular*}{5.8in}{@{\extracolsep{\fill}}p{2.2in}p{0.3in}p{2in}p{0.3in}}
    \toprule    
    \multicolumn{4}{ c }{\textbf{Characteristics of Study or Research Designs}} \\ \midrule
    \centering\textbf{Characteristics} & {Count} & \centering\textbf{Characteristics} & {Count}\\
    \cmidrule{1-2}
    \cmidrule{3-4}
    \centering \textbf{Benchmark Datasets used} &   & \centering \textbf{Languages of datasets} $^\ddag$ &   \\
    \cmidrule{1-2}
    \cmidrule{3-4}
    \centering \footnotesize \citet{waseem-hovy-2016, waseem-2016} & \quad \footnotesize 14 & \centering \footnotesize English & \quad \footnotesize 36 \\ 
    \centering \footnotesize \citet{fersini_2018, evalita_2020} & \quad \footnotesize 13 & \centering \footnotesize Spanish &  \quad \footnotesize 7 \\
    \centering \footnotesize \citet{basile-etal-2019} & \quad \footnotesize 5 & \centering \footnotesize Italian & \quad \footnotesize 6 \\
    \centering \footnotesize \citet{bates_2015} & \quad \footnotesize 6 & \centering \footnotesize South Asian languages (e.g., Bangla, Hindi) & \quad \footnotesize 6 \\
    \centering \footnotesize \citet{rodriguez_2021}  & \quad \footnotesize 4 & \centering \footnotesize Other European languages & \quad \footnotesize 2 \\
    \cmidrule{3-4}
    \centering \footnotesize \citet{fersini_2018}  & \quad \footnotesize 3  & \centering \textbf{Paradigms} \footnote{More on the paradigms \citep{röttger_2022} in the Supplementary Section \ref{table:terminologies}, in the table of terminologies.} $^\ast$$^\ddag$ &  \\
    \cmidrule{3-4}
    \cmidrule{1-2}
    \centering \textbf{Machine Learning models} &  & \centering \footnotesize Perspective & \quad \footnotesize 37 \\
    \cmidrule{1-2}
    \centering \footnotesize Support Vector Machine (SVM) & \quad \footnotesize 16 & \centering \footnotesize Descriptive & \quad \footnotesize 5 \\
    \centering \footnotesize Bidirectional Encoder Representations from Transformers (BERT) & \quad \footnotesize 15 & \centering \footnotesize Unsupervised (as per model features) & \quad \footnotesize 3 \\
    \cmidrule{3-4}
    \centering \footnotesize Long Short-Term Memory (LSTM) & \quad \footnotesize 15 & \centering \textbf{Evaluation type} $^\ddag$ &  \\
    \cmidrule{3-4}
    \centering \footnotesize Logistic Regression (LR) & \quad \footnotesize 10 & \centering \footnotesize Binary Classification & \quad \footnotesize 32 \\
    \centering \footnotesize Convolutional Neural Networks (CNN) & \quad \footnotesize 9 & \centering \footnotesize Multi-class Classification & \quad \footnotesize 22 \\
    \centering \footnotesize Naive Bayes (NB) & \quad \footnotesize 6 & \centering \footnotesize Multi-label Classification & \quad \footnotesize 4 \\
    \centering \footnotesize Random Forest (RF) & \quad \footnotesize 6 & \centering \footnotesize Cluster Analysis & \quad \footnotesize 1\\
    \centering \footnotesize Decision Tree (DT) & \quad \footnotesize 4 & \centering \footnotesize Results per class & \quad \footnotesize 1\\
    \cmidrule{3-4}
    \centering \footnotesize Multilayer Perceptron (MLP) & \quad \footnotesize 4 & \centering \textbf{Annotator types} &  \\
    \cmidrule{3-4}    
    \centering \footnotesize XGBoost & \quad \footnotesize 3 & \centering \footnotesize External dataset & \quad \footnotesize 24 \\
    \centering \footnotesize fastText & \quad \footnotesize 2 & \centering \footnotesize Experts & \quad \footnotesize 5\\
    \cmidrule{1-2}
    \centering \textbf{Platform of interest} &  & \centering \footnotesize Authors & \quad \footnotesize 4\\
    \cmidrule{1-2}
    \centering \footnotesize Twitter (or X) & \quad \footnotesize 28 & \centering \footnotesize Amateurs/Crowdsourced external annotators & \quad \footnotesize 3\\
    \centering \footnotesize Sexism reported online (from Everyday Sexism Project) \citep{bates_2015} & \quad \footnotesize 3 & \centering \footnotesize Students of linguistic, communication and gender & \quad \footnotesize 3\\
    \centering \footnotesize Facebook & \quad \footnotesize 2 & \centering \footnotesize Machine learning models & \quad \footnotesize 3\\
    \centering \footnotesize Reddit & \quad \footnotesize 2 & \centering \footnotesize Social Scientists & \quad \footnotesize 2\\
    \centering \footnotesize Gab & \quad \footnotesize 2 & \centering \footnotesize Annotator character not stated  & \quad \footnotesize 2\\
    \bottomrule
    \end{tabular*}
    \caption{\centering Summary table of some of the topmost categories of designs/methodologies in all \hspace{\textwidth} the observed characteristics across the selected studies. \hspace{\textwidth} \footnotesize $^\ast$ indicates unique features of a document, i.e., a document can only have either of the categories in the field. \hspace{\textwidth} $^\ddag$ indicates that the categories listed below are a part of the exhaustive list for that particular field.}\label{table:sys_obs}
\end{table}
\raggedbottom

\subsubsection{Overview of the general methodologies}
\begin{sidewaysfigure}
\centering
 \includegraphics[width=\textheight, keepaspectratio]{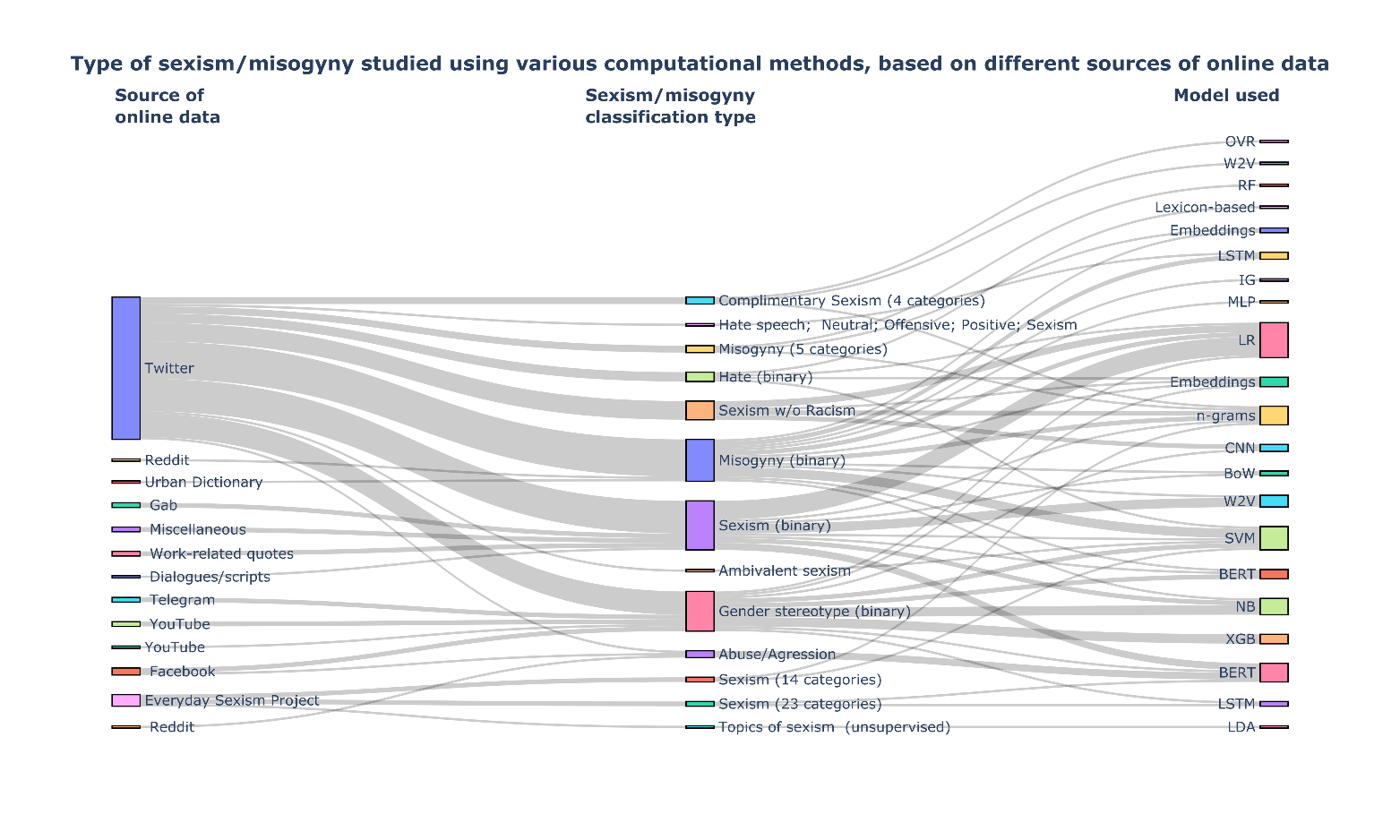}
 \caption{Sankey diagram of the link between each categories of online data, classification type and computational models. The Sankey Diagram allows to visualize flow between various points in a system. In our system, we show the flow (i.e., count of association) between sources of online data, sexism/misogyny classification type, and NLP model used.}
 \label{fig:sankey}
\end{sidewaysfigure}

As per the Table \ref{table:sys_obs}, we do see the frequency of each source of online data and the machine learning models used in the 45 manuscripts we reviewed. The type of classification of sexism and misogyny used in the said studies are otherwise unknown and how they link between the sources of online data and the computational methodologies is an important source of information to indicate the multi-level connections between the variables, and consequently its impact on the quantification of sexism and misogyny. Figure \ref{fig:sankey} show the connection between nodes in each level of information (source of online data, classification type and model used), while the connections are the links between each level with their weights indicating the frequency of connections between each node type at different levels. This is a many-to-many mapping between the three levels, and show the flow of information between each of them. The colors of each node at different levels represent the unique relation between the linked nodes at each level (e.g., Twitter [level 1] $=>$ Misogyny (5 categories) [level 2] is different from Twitter [level 1] $=>$ Misogyny (binary) [level 2], and Reddit [level 1] $=>$ Misogyny (binary) [level 2] is different from Reddit [level 1] $=>$ Abuse/Aggression [level 2]). The abbreviations for the models in level 3 are found in Supplementary Section \ref{abbr:models}. Overall, the Figure \ref{fig:sankey} gives a clear evidence that Twitter was the most explored online data source to investigate different forms of sexism and misogyny.

\subsection{Overview of the existing computational approaches}

\subsubsection{General definitions and strategies used}
Most computational works on sexism and misogyny took the automated identification problem as a binary classification task, i.e., deciding if the text in question is sexist/misogynistic or not. For defining the terms, researchers used different (non-standardized) forms for their work because of computational benefits, such as model performances. Though they served to be effective in some instances, they also presented with limitations of their own. For instance, \citet{grosz_conde_2020} define sexism as the prejudicial and discriminatory nature of sexist behavior pervading in the social context, especially for women. Using theoretical concepts, along with the typology of abuse presented in earlier research, \citet{guest-etal-2021} define misogyny as content “directed abuse at women or a closely-related gendered group (e.g., feminists).”  Whereas \citet{lynn_endo_2019} define misogyny as a hate crime, a result of a “cultural attitude of hatred for females because they are female”, and presented with two bidirectional-DL models (Bi-LSTM and Bi-GRU) with dropout layers, which performed well in sensitivity and accuracy even with a slightly imbalanced dataset.

Beyond binary classification, combining techniques can enhance model performance. For example, \citet{attanasio_2020} demonstrated improved results using a multi-agent classifier built upon sentence embeddings, TF-IDF weighting, and lexicons specific to misogynistic language.\citet{plaza_2021} proposed Multi-Task Learning (MTL) system with hard parameter sharing approach (sharing the hidden layers between all tasks, while keeping several task-specific output layers) using BERT-based models for utilizing the transferred knowledge from multiple other (but related to sexism identification) tasks like polarity and emotion classification and offensive language detection classification helped in the identification, both in binary and multiple categories. Though emotion was not helpful in categorizing, MTL shows promising generalization to the original task.

Additionally, \citet{frenda_2019} exploit stylistic, semantic and topic information about misogynistic speech to identity misogyny and classify it to different categories. For gathering linguistic features, they propose an approach based on stylistic features captured by means of character n-grams, on sentiment information and on a set of lexicons built by examining the misogynistic tweets from training data provided by the organizers. Each text was represented by a vector composed of all specific topic features(set of lexicons), pondered with Information Gain, and character n-grams, weighted with TF-IDF measure. This set of features were experimented employing a Support Vector Machine (SVM) algorithm and an ensemble technique, reaching promising results. \citet{cans_2018} worked on the same data and task, experimenting SVM alongside TF-IDF with a one-vs-one and one-vs-rest classifier approach, where the later proved better for English, presumably because of larger vocabulary. On the other hand, \citet{nozza_volpetti_2019} defined several templates to create a balanced synthetic dataset for their proposed DL model- Universal Sentence Encoder (USE), which further debiased their model features to be less-sensitive to identity terms and yet obtain a better categorization.

\subsubsection{Overview of performance evaluation}
When it comes to evaluating performance, classification models with traditional computational approaches seem to fair relatively similar or better (in some cases) at automated sexism/misogyny identification tasks. \citet{indurthi_etal_2019}'s work shows how different set of pretrained embeddings trained from different state-of-the-art architectures and methods when used with simple machine learning (ML) classifiers like SVM and XGBoost perform very well in binary classification tasks. \citet{kohli_etal_2021} used two kinds of methods: first using an ensemble approach comprising of XGBoost, LightGBM and Naïve Bayes; and second employing BERT-based architecture. Both the models performed well on binary identification task, but differently on different languages and aggression label analysis, one of which was gendered, due to the overlapping context in all. Overall, SVM is seen to be the best-performing conventional classifier (hence taken as a baseline for some of the works), and a lot of papers have used it in their work as a standalone classifier, or as an ensemble voting classifier \citep{frenda_2018, francimaria_2022}- alongside other classifiers like Gradient Boosting and Random Forest (RF). Regression models like Logistic Regression have also been used by a lot of studies, especially for binary tasks.  While Decision Tree \citep{plaza_del_arco_2020}, and RF \citep{singh-etal-2021} has also been used, they do not show much success among the conventional ones whereas most DL models use Fully Connected (FC) layers for classification \citep{bashar_2020}. Yet Convolutional Neural Network (CNN) with various semantic features is also seen to work well in categorizing hate-speech -- particularly sexism, even outperforming Logistic Regression for some of the performance metrics \citep{gamback-sikdar-2017-using}. 

However, there is no definitive indication of whether traditional models or neural networks (DL models)—particularly transformer-based architectures—are generally preferred for detecting or identifying sexism or misogyny. While some studies did not perform comparative analyses across different model architectures (e.g., \citep{waseem-2016, shah-etal-2020, frenda_2018, cans_2018}), others explicitly compared selected models to highlight performance differences. For instance, some studies such as \citet{plaza_del_arco_2020, schütz_2022, butt_2021} found that traditional models (e.g., SVM, Random Forest) outperformed neural networks, whereas others like \citet{sharifirad_2018, lynn_endo_2019, guest-etal-2021} reported the opposite. Additionally, some studies showed comparable results across different model types (e.g., \citep{grosz_conde_2020, singh-etal-2021, bashar_2020}). Transformer models have been shown to offer clear advantages over conventional neural network models in several studies (e.g., \citep{chiril-etal-2021, chiril_etal_2020, parikh_2019}). Learning sequence and contextual information through embeddings yielded competitive performance in certain cases, particularly using LSTM and BERT-based models (e.g., \citep{Badjatiya_2017, rodriguez_2020, chiril-etal-2021}). More broadly, the integration of embeddings was generally found to enhance model performance (e.g., \citep{fersini2021deep, anzovino_2018, jha-mamidi-2017}). Furthermore, incorporating text features such as n-grams (e.g., \citep{bhattacharya_etal_2020, waseem-hovy-2016, butt_2021}) or combining both embeddings and n-gram features (e.g., \citep{abburi_2021, sharifirad_2018}) also contributed to improved outcomes.

Yet, studies such as \citet{grohndal_2018, Arango_2022} suggest that most research emphasizes model development, with limited attention to evaluating whether these models generalize to other contexts or understanding the factors that influence their success or failure. In fact, \citet{Arango_2022} found a \emph{significant drop} in performance while replicating some state-of-the-art methods across multiple datasets. \citet{herd_2024} explicitly state that \emph{``relying on such single point estimates to evaluate safety requirements is problematic since they only provide a partial and indirect evaluation of the true safety risk associated with the model and its potential errors''}. Given the probabilistic result the metrics provide over a binary domain, it may overestimate the model performance since their trustworthiness depends on other secondary factors such as sample size, the model calibration, the quality of the dataset, among others \citep{herd_2024}.
This raises the question of whether strong performance metrics alone are truly reliable indicators of a robust or effective model.

\subsubsection{Is classification the only way?}
Though almost all the computational methods employ classification techniques, it is not the only way. But it is favorable, for good reasons. Clustering techniques are mostly useful for content analysis and to study discourse, to help identify implicit themes/topics from the data which may be (unintentionally) omitted during manual inspection, and reassignment into its overarching categories for better interpretation, even though may sometimes provide superficial results \citep{siddiqi_2018}. \citet{karami_2019} employs unsupervised text-mining approaches like LDA topic modeling. Utilizing the themes they found, they performed qualitative thematic analysis before finally moving to a theoretical thematic analysis to group the previously identified topics into four categories of sexism. \citet{melville_2019} also uses LDA for grouping 7 topics, and alongside clustering based on Louvain algorithm \citep{blondel_2008} for grouping 20 topics. They define sexism based on themes and sites associated with the experience of sexism from Everyday Sexism \citep{bates_2015} and journalism. From these studies, it is evident that clustering is more useful for content analysis, rather than for the detection/identification tasks.

\subsection{Challenges}\label{sec:challenges}

\noindent In this section, we outline the challenges for the interdisciplinary approaches that are likely due to the disciplinary divide, and ways of addressing them.
We identify that these challenges could be because of two broad reasons - (i) Use of different computational strategies; and (ii) Linking social science theories to the tested computational strategies. The first part essentially talks about the different strategies used, compares them based on different parameters in each subsection, and weighs the advantages and limitations of each approach. The second part focuses more on how existing literature has sexism and misogyny in their work. By analyzing how the same terms are defined in social science theories, we form an argument about how the existing computational research could benefit from a more fine-grained categorization of sexism (or misogyny) to improve their automated identification task. 

\subsubsection{Use of different computational strategies}
This subsection is intended to shed light on the different computational strategies that have been used to quantify sexism and misogyny, while also segregating the strategies based on some differentiators like the small dataset size and the dataset languages used.

\paragraph*{Beyond Binary Classification}

Most of the studies have used binary classification of sexism (e.g., \citep{singh-etal-2021, schütz_2022, grosz_conde_2020}) and misogyny (e.g., \citep{bashar_2020, lynn_endo_2019, bashar_2020}) for their detection tasks. While practical, such models often overlook theoretical nuance, providing surface-level predictions that may fail to capture the complex manifestations of sexism \citep{parikh_2019, abburi_2021}. Previous studies have highlighted the difficulty in capturing the nuanced meanings of terms like sexism and misogyny in computational settings \citep{samory_sen_2021}, emphasizing the importance of evaluating both the intensity and specific type of misogynistic behavior within a given context \citep{lynn_endo_2019}.

Tasks such as hate speech or harassment detection inherently involve subjectivity, with no universally accepted definitions or absolute truths \citep{röttger_2022}. While certain beliefs may be broadly agreed upon, they do not fully encapsulate the complexity of these concepts. Moreover, annotation processes are often conducted without direct oversight from dataset creators, resulting in partially subjective datasets that may not align clearly with specific downstream applications \citep{röttger_2022}. Even with detailed label descriptions, human annotators often struggle with intuitively unclear \citep{vidgen_2020} or closely related \citep{parikh_2019} categories. Some improvements have been observed when annotation guidelines are adjusted to address edge cases \citep{zeinert-etal-2021},though further distinction—such as between active and passive language—may offer additional clarity \citep{anzovino_2018}. Moreover, perceptions of toxicity are highly subjective and often shaped by the receiver's interpretation of the speaker’s expression \citep{sap-etal-2019}. 

Despite increasing interest, research addressing the nuanced forms of sexism and the intersectional ways women experience it—online and offline—remains limited \citep{melville_2019}. \citet{barak_2005} identifies four types of gender-based harassment in cyberspace: active verbal, passive verbal, active graphic, and passive graphic. These categories are influenced by both objective features (e.g., explicitness, repetition) and subjective factors (e.g., recipient's attitudes, sensitivities). As a result, perceived severity varies widely across individuals—a principle equally applicable to sexism and misogyny. 
Similarly, \citet{sharifirad_2018} propose a four-type classification of sexism: information threat, and indirect, sexual, and physical harassment. To overcome data scarcity, they use ConceptNet to semantically enrich text via augmentation techniques—leveraging relations like “IsA” and “RelatedTo”—with three replacement strategies: all words (most effective), nouns, and verbs. They find that such semantic enrichment, especially through text generation alone, considerably boosts classification performance.

\citet{swim_mallett_2004} point out that using only sexist language—as done in the above studies—limits the generalizability of findings to other, less overt forms of sexist behavior. Still, multi-label classification has shown promise. Echoing this approach, \citet{talavera_2021} also employ a multi-label model with fewer (five) categories. They reinforce earlier findings that pretrained language models, when fine-tuned to the specific task domain, are particularly effective in low-resource settings.
Most of the existing research on multi-class classification of sexism or misogyny consider at most five categories of sexism (such as \citep{anzovino_2018, sharifirad_2018, jha-mamidi-2017}). \citet{anzovino_2018} classify misogyny into five subcategories: Discredit, Harassment \& Threats of Violence, Derailing, Stereotype \& Objectification, and Dominance. \citet{guest-etal-2021} extend from their work to develop a hierarchical taxonomy with three levels to define misogynistic content, with four overarching categories of misogyny: (i) Misogynistic Pejoratives, (ii) descriptions of Misogynistic Treatment, (iii) acts of Misogynistic Derogation and (iv) Gendered Personal attacks against women. Studies like \citet{frenda_2018, cans_2018, shah-etal-2020} use the IberEval 2018 shared task dataset \citep{fersini_2018} where the data was categorized to five categories, namely: \emph{Stereotyping and Objectification}, which involve oversimplified portrayals of women or emphasis on their physical appearance; \emph{Dominance}, which asserts male superiority to reinforce gender inequality; \emph{Derailing}, which shifts blame away from men or disrupts conversations to refocus them around male comfort; \emph{Sexual Harassment and Threats of Violence}, which encompass unwanted advances, coercion, or threats intended to exert power over women; and \emph{Discredit}, which targets women with slurs or insults without any broader argumentative context or intent.

Likewise, the EXIST shared tasks in 2021 and 2022 \citep{rodriguez_2021, rodriguez_2022}  begin with a binary definition of sexism based on the Oxford English Dictionary — ``prejudice, stereotyping, or discrimination, typically against women, based on sex'' — and extend it into a multi-class taxonomy comprising five expert-defined categories: \emph{Ideological and inequality}, which includes anti-feminist rhetoric or denial of structural gender inequality; \emph{Stereotyping and dominance}, referring to assertions of traditional gender roles or male superiority; \emph{Objectification}, where women are reduced to physical appearance or sexualized attributes; \emph{Sexual violence}, encompassing harassment, coercion, or threats of a sexual nature; and \emph{Misogyny and non-sexual violence}, involving expressions of hatred, hostility, or aggression directed at women. Both datasets support the detection of a broad spectrum of sexist content, ranging from explicit misogyny to more subtle, implicit expressions of sexist behavior, as opposed to most other research which only focus on the detection of explicit contents (such as \citep{waseem-hovy-2016, katsarou_2021, anzovino_2018}). 

The utility of multi-label classification is highlighted in \citet{parikh_2019}, who illustrate the co-occurrence of various sexism categories in user-reported first-person accounts. Their dataset comprises 23 categories (defined by a social scientist) derived from real-world reports of sexism. The annotation process followed a rigorous three-phase protocol involving pre-training, pilot testing, and final quality checks, ultimately reducing the labels to 14 for classification. To address domain-specific challenges, they fine-tuned a BERT model using masked language modeling and next sentence prediction, further enhancing it with distributional word embeddings and a linguistic feature vector. These linguistic features drew from prior work on biased language detection and incorporated affective and sentiment lexicons, including PERMA (Positive Emotion, Engagement, Relationships, Meaning, Accomplishment) features and emotion scores. Their proposed multi-label, multi-class LSTM-based framework significantly outperformed various traditional machine learning and deep learning baselines. 

Building on this work, \citet{abburi_2021} take a more fine-grained approach by reintroducing the full 23-category schema. Using self-trained semi-supervised learning, they augment the labeled dataset to allow co-occurrence of categories and improve class distribution. Their approach enhances textual diversity and focuses on hard-to-classify samples using confidence scores and selective intersection. In addition to the domain-tuned BERT with a BiLSTM and attention mechanism, they incorporate a custom loss function that leverages label confidence scores. Their domain-tuned BERT–BiLSTM–Attention model achieved notable gains, though the data focused exclusively on sexist examples and did not generalize to broader detection tasks.

\paragraph*{Data Limitations and Model Performance}

Even when models perform well, limited dataset size can undermine reliability and restrict exploration of more sophisticated deep learning (DL) architectures, such as those employing advanced attention mechanisms requiring larger amounts of data \citep[e.g.,][]{grosz_conde_2020}. For example, in their binary classification of misogyny, \citet{guest-etal-2021} developed a hierarchical taxonomy across three levels: the second level featured four non-mutually exclusive categories—misogynistic pejoratives, treatment, derogation, and gendered personal attacks—with further sub-categorization at the third level. Although the classification is sound, and logistic regression and both weighted and unweighted BERT achieved strong performance, the dataset's limited size and low proportion of misogynistic content (8.1\%) posed challenges, especially since classification was performed only on the instances labeled True for misogyny.

To address small dataset issues, \citet{schütz_2022} explored transfer learning approaches, including the use of multilingual transformers pre-trained on external datasets and data augmentation by integrating similar external content. Their experiments demonstrated that fine-tuning the entire model with domain-specific data improved performance. However, pre-training generally proved more effective, as fine-tuning showed tendencies toward overfitting and did not yield consistent improvements on external datasets. As  DL models typically require large datasets to achieve optimal performance, traditional models may give a comparative performance (such as \citep{samory_sen_2021}) or even outperform them in data-constrained settings -- particularly when augmented with rich linguistic features such as n-grams \citep{plaza_del_arco_2020, butt_2021}.

\paragraph*{Dependence on external benchmark datasets}

Several previous studies highlight that the main challenge in quantifying sexism stems from the lack of high-quality datasets required to train robust, scalable automated detection systems \citep{guest-etal-2021}. Most computational studies rely heavily on external benchmark datasets, which can affect the quality, reliability, and representativeness of the data. As \citet{zeinert-etal-2021} point out, “When abusive language is annotated, classes are often created based on each unique dataset (a purely inductive approach), rather than leveraging established terminology from social science or psychology (a deductive approach, building on existing research).”

While benchmark datasets are frequently used in shared tasks\footnote{“Shared tasks are collaborative efforts in which researchers and practitioners come together to solve a common problem using shared data and evaluation measures. They promote competition, collaboration, and progress in research, and have become an important part of many academic and industrial communities” \citep{SIGEDU_2024}.} for automated misogyny detection and related challenges, they are often found to be misrepresented or imbalanced. For example, the IberEval2018 dataset contained uneven category representation—certain misogynistic behaviors, like ‘derailing,’ were underrepresented ($<$2\%), while categories like ‘active’ tweets were significantly overrepresented ($>$85\%). Additionally, notable discrepancies appeared between language groups in the same dataset \citep{cans_2018}. Similarly, in the TRAC2020 shared task dataset \citep{trac2-dataset}, texts containing multiple languages were categorized under a single language, complicating analysis for non-speakers unfamiliar with the other languages' socio-cultural contexts. Furthermore, the proportion of texts exhibiting hate speech varied significantly across languages \citep{gordeev-lykova-2020}.

These inconsistencies underscore the necessity for dedicated, theoretically-informed, ‘reliable’ data collection and annotation practices grounded in social science frameworks, before performing any experimentation with NLP tools. 

\paragraph*{Datasets are mostly in Western Languages}
The majority of studies on online hate speech detection have been conducted in English (e.g., \citet{katsarou_2021, waseem-hovy-2016, parikh_2019}, \emph{inter alia}). Spanish and Italian follow closely, particularly through shared tasks—for example, studies in Spanish \citep{schütz_2022, plaza_del_arco_2020, cans_2018} and in Italian \citep{attanasio_2020, ou_li_2020, nozza_volpetti_2019}, \emph{inter alia}. More recently, a limited number of works have begun to explore other languages such as Hindi, Bangla (e.g., \citep{bhattacharya_etal_2020}) and Chinese (e.g., \citep{jiang2021swsrchinesedatasetlexicon}).
But even within same languages, there lies substantive differences in the peculiar lexical choice and morphological structures rising from the regional colloquial usages, leading to linguistic and cultural heterogeneity \citep{bhattacharya_etal_2020}. However, this fundamental issue remains largely unexamined in the current literature. Though models like fine-tuned cross-lingual multitask BERT shows promising performance even with non-English languages, with it performing better on Bangla when experimented alongside English, presumably owing to the dataset peculiarity or specific features of the language itself \citep{gordeev-lykova-2020}. But when it comes to representation, there is a huge gap between only the use of Indo-European languages (especially English) and other languages. When working with multiple languages, “back-translation” has been used in the said languages to augment the data and translate all of them to a single uniform language, which could be one of the source languages or different. 

\citet{butt_2021} performed the same technique on Spanish and English (source) languages to convert it into English, with German being the second language, using the deep-translator python library. And their results on all of their tried ML algorithms show an improvement with the augmentation; even indicating that with proper pre-processing, it could give competitive results in comparison to deep learning models. \citet{zeinert-etal-2021} too had experimented translating misogynistic posts provided by \citet{anzovino_2018} to Danish using translation services in an attempt to augment the minority class data. But it did not prove as useful in providing a sampling alternative, hence inferring that language-specific investigation is important for cultural discovery, for the sake of automatic detection systems. \citet{rodriguez_2020} found improved cross-domain generalization when models were trained on their Spanish dataset and tested on an Italian dataset. This suggests that some datasets may offer broader coverage of sexist content—including explicit, implicit, and context-dependent forms—and are therefore more transferable for studying the same task across different languages. However, we observe a lack of experiments involving languages that differ significantly in cultural-linguistic context and core linguistic dimensions.

In fact, \citet{waseem-2016} suggests against boosting the minority class in the interest of mimicking reality in the datasets, even if it causes larger misclassification for the class. \citet{rahali_akhloufi_2021} uses gender swap data augmentation and data consolidation with feature ablation, which is seen to improve the learning of the model, especially when used with the same language. But using multi-language datasets does not help much, since English does not consolidate well with other languages (e.g., Arabic and French) with limited samples as compared to English, inevitably giving rise to data imbalance, a data bias. So, there is a need to look beyond English. 

\citet{singh-etal-2021} converted the whole dataset from multiple languages to a uniform English dataset, by transliterating the sentences belonging to other languages using IndicTransliterator. But that required transferring a word from the alphabet of one language to another, which could give faulty outcomes. Given the linguistic variety and limitations that could be faced when delving into other languages, improving the existing gaps in sexism identification tasks in English should be of primary focus.

\paragraph*{Biases }
Bias is a broad term that can be defined in various ways, depending on the field and context. In this section, we summarize: (a) how studies in our dataset define or quantify biases, (b) whether these studies acknowledge biases present in their own research, and (c) what measures, if any, they adopt to mitigate these biases. Social science studies tend to address bias in their work. In NLP, however, definitions of bias often depend heavily on domain-specific concerns like biases in word embeddings, annotator labels, or amplified predictions related to demographic characteristics \citep{hovy_2021}. In this work, we adopt the definition by \citet{shah-etal-2020}, where bias refers specifically to “the mismatch of ideal and actual distributions of labels and user attributes in the training and application of a system.” Additionally, the rapid advancement of NLP may itself exacerbate bias by outpacing the field's ability to adapt effectively to emerging contexts \citep{hovy_2021}. \\

\noindent\textbf{Gender bias}\\
\noindent\citet{eagly_mladinic_1989} use attitude theory to explain how stereotypes emerge from attitudes toward men and women. According to them, the cognitive component of an attitude --- defined as a person's thoughts about an object --- can manifest through attributes assigned to social groups. When attributes assigned to these groups carry evaluative meanings (positive or negative), they give rise to stereotypical perceptions. Their study reveals that even when women are positively evaluated, they tend to be stereotypically viewed as inferior to men in agentic or instrumental (traditionally masculine-positive) qualities, yet superior in communal or expressive (traditionally feminine-positive) qualities.

Further supporting this, \citet{schmid_2004} provide empirical evidence of an implicit hierarchical gender stereotype, showing that men are associated more strongly with hierarchy and women with egalitarianism. The magnitude of this stereotype reflects a deeply rooted societal bias. Social media platforms further amplify these biases through their widespread influence. Consequently, biased social information from such sources, when used to train machine learning models, can unintentionally lead to gender biases. Models may thus propagate negative stereotypes about various social groups.

To address this issue, \citet{dinan_fan_et_al_2020} proposed measuring bias through a semantic and pragmatic framework, evaluating three dimensions derived from “knowledge of the conversational and performative aspects of gender.” By separately analyzing how author gender influences the dataset, they aimed for a deeper understanding and mitigation of gender bias.  \\ 


\noindent\textbf{Annotator bias:}\\
\noindent It is seen that having a lot of categories of misogyny may also impact on the annotators’ agreement, both in terms of depth (subcategories) and breadth (different types) of the said categories, owing to the differences in experience and values of the annotators. And their inherent social biases may impact on their choice, especially when working using contexts \citep{guest-etal-2021}. Having different level of understanding of the language in question or personal prejudices, and differing individual world-view are seen as primary issues in inter-annotator disagreements. \citet{bhattacharya_etal_2020} used several rounds of discussions and sensitization towards gender issues among annotators to resolve this issue, by providing with counterexample method and examining annotator votes, alongside using an ‘unclear’ tag in case of disagreement. Sometimes when adversarial examples (even just 25\%) are included while training the dataset, it is seen to help in the robustness of the models and their performance. In fact, providing the models with different aspects of sexism and challenging the models with different examples have shown to be effective for generalizability \citep{samory_sen_2021}. 

Some studies use different criteria for selecting the annotators they want in their study, based on both similarities and differences on each. It could be based on region, demographic, education and ethnicity \citep{guest-etal-2021}, native speakers of language \citep{chiril_etal_2020}, feminists \citep{jha-mamidi-2017}; but mostly who studied gender \citep{lynn_endo_2019} and linguistics \citep{nozza_volpetti_2019}. A comparison on amateur (crowd-sourced workers) and expert (having  both  a  theoretical  and  applied  knowledge of hate speech) annotators (as most studies use either of them) by \citet{waseem-2016} state the contrasts observed in annotation with both, and the consequent model performances which did not substantially improve on their previous model \citep{waseem-hovy-2016}. The emphasis on the most significant features changes from extra-linguistic features for majority-voted amateurs to content of the tweets for the experts, and among the features they experimented with, the ones having highest performances (high F1 score) were not necessarily the features with the best performances. \citet{singh-etal-2021} considers misogyny and sexism as a subset of hate-speech, and used data that was manually annotated by multiple annotators using 'Discursive Methods of Annotation' since it was seen as a pragmatic approach to including the socio-pragmatic phenomenon using social studies, and as a function of both the contextual factors and the discursive experience of the speaker. \citet{zeinert-etal-2021} does an iterative process of raising cases for revision in the discussion rounds, formulating the issue, and providing documentation for annotation, inviting in annotators with diversity in age, occupation/background, region (spoken dialects).
Annotation biases can lead to other kinds of bias, like racial bias due to lack of knowledge of different dialects- which could potentially amplify the harm against people from the minority community \citep{sap-etal-2019}. \\

\noindent\textbf{Other biases}\\
\citet{hirsch_1992} documents the oppression of women through language, as she talks about male-specific words that are positively portrayed in English, in turn reflecting the “consensus reality” of the patriarchal society. While theorizing the language and gender connection (with many of the examples drawn from political discourse) from one of its reviewed books, it talks about how language is used as a tool to further perpetuate patriarchy \citep{cameron2020language}.  The same is seen for the computational models based in English. The datasets taken for studies could also add to the bias owing to the different considerations made due to the data source, hence not representing the diversity in real-world. For example, domain sources where misogyny is assumed to be most likely like women fashion blogs, fitness tips videos, etc. \citep{bhattacharya_etal_2020}; or when sexism is taken as one of the sentiment label, with data collected around some specific cases/instances/event networks like \#Coronavirus, \#ClimateChange, \#Immigrants and \#MeToo \citep{katsarou_2021}. Another form of such bias could rise because of subjectivity in mislabeled data. \citet{samory_sen_2021} had performed re-annotation on the external datasets they used in their study, following the sexism annotation codebook they devised themselves. Relying on two baselines: Gender-Word \citep{zhao-etal-2018} and Jigsaw’s Perspective API \citep{hosseini_2017}, they found a large majority of sexist tweets were non-sexist, only $\sim$60\% of the sexist labels adhering to their ground truth. They found that stratifying misclassification rates helped in giving a more accurate result. Both these points could hinder model performance.

\hfill\break
Yet, with the listed biases, always a question remains if they were a cause of systematic errors (both conscious or unconscious) or were a result of a narrowed preference in a particular direction in favor of the said bias. In other words, the use of `bias' to refer to systematic error is problematic. According to \citet{hammersley_gomm_1997}, it depends on `truth' and `objectivity', whose justification and role have been questioned. Due to the ambiguous nature of the term itself, we might question if the forms of bias explored are a result of methodological adaptability; conscious limitations due to the scope of the research (such as research designs); or could arise because of the models themselves. Either way, they may not indicate the research as ``being biased''. Of the five most common sources of bias in NLP tasks as identified by \citet{hovy_2021}, we have reviewed almost all of them in this section. This indicates that these biases are well-known across the CSS literature, and can be explored more to mitigate them from all sources, using algorithmic and methodological approaches.

\paragraph*{Lexical dependency for characterizing sexism and misogyny}

To linguistically characterize misogyny and sexism, many studies have used different theoretical concepts to represent both. \citet{farrell_fernandez_2019} had built a list of key lexicons for categorizing misogyny using Encyclopaedia of Feminist Theories \citep{code_2002} and other pre-existing hate-speech lexicons and studies of the specific rhetoric of manosphere, taken from different corpus. In their observatory work, they study the evolution of communities where users share in-group characteristics. But even though corroborating the theories and existing ideas helped in providing lexicons, they acknowledged the limitations of using it due to its lack of completeness (shortcomings in capturing all the words that might be relevant). Other times, studies use words ‘typically associated’ with misogynistic content created by domain experts \citep{lynn_endo_2019} which is used as neologisms for identification of emerging or cloaked misogyny. \\

Lexical dependency can cause NLP models to overfit because of too much influence of certain identity terms. This eventually results in false positives, severe unintended bias, and lower performance. \citet{rodriguez_2020} acknowledge the biases inherent in keyword-based dataset collection, noting that models tend to exhibit bias toward specific terms and struggle to detect instances featuring shorter length, subtle expressions of sexism, irony, or context-dependent language.
\citet{bashar_2020} acknowledge that misogynistic abusive tweets might contain certain keywords, but would not necessarily always contain such slurs. To work around that, they show that classifiers can work with small-labeled datasets, provided that the word vectors used are pre-trained on the context domain of the problem and paired with careful customization and regularization. This proves that a large-labeled dataset is not always required for training purposes. In fact, if the word vectors are pre-trained in the context of the problem domain, alongside careful customization of the model, the classifiers could also be trained on small datasets. On the other hand, \citet{plaza_2021} generates linguistic resources using a set of word embeddings, with the initial seed lexicon eventually getting populated with words and n-grams more attuned to the domain because of linguistic similarities. Using a voting schema rule with logistic regression and multinomial Naïve Bayes, alongside the lexicon-based system and combinations of unigrams and bigrams gave a good result with the Spanish dataset. Observations show that some expressions of hate when combined with other terms change the sense entirely and hence better-supervised learning begins with larger data. \\

For larger datasets, the issue is elevated with the imbalanced nature of the datasets and their disproportionate dependence on these determinate terms, having a high correlation to minority class \citep{nascimento_2022}. Using such identity terms, or samples from target domains during the training phase requires a-priori knowledge but can often lead to the introduction of further bias. Introducing a regularization approach to the models to add some degrees of contextualization using Entropy-based Attention Regularization (EAR) could mitigate the problem to some extent, as they are seen to show competitive performance, along with an improvement in the bias metrics \citep{attanasio_2022}. Consequently, developing classifiers that can decompose gender bias within full sentences into semantic dimensions can be used, since it can be contextually determined (rather than being explicitly gendered). This has in turn shown to give a better performance in controlling gender differences \citep{dinan_fan_et_al_2020}. \citet{ou_li_2020} find limitations of only using the pooler output of DL multilanguage models like XLM-RoBERTa, and hence obtains deeper and more abundant semantic features by extracting from its hidden layer state which gives better performance. Data correction strategy focused on gender bias, consisting of two-stage modules- bias detection and replacement of the said bias-sensitive words (BSWs), is seen to reduce the differentiation of similar terms related to gender, and in turn, contribute to mitigating the unintended bias. Since the frequency of female identity terms is high (even when representing similar groups/classes or other social identities) in datasets related to sexism and misogyny, they replaced these potential bias terms with $<$identity$>$ tag without compromising the model accuracy. Their proposed multi-view stacked classifier is seen to outperform other state-of-the-art models and diminish gender bias \citep{nascimento_2022}. 

Natural language processing (NLP) applications like sentiment analysis are crucial for analyzing and detecting online sexism/misogyny. Incorporating polarity and emotion information is seen to be useful for the benefit of the task as they portray the usually emotional, expression of negative emotion and polarity towards the recipient \citep{plaza_del_arco_2020, plaza_2021}. Using feature representations has further helped in training the model, by adding representations of the text in terms of various lexical, syntactic, and morphological features. While the most common types of features used are the bag-of-words representations of text, and/or the embeddings, adding to the features also helps in the performance. Many papers have used it to enhance their model performance. The idea is to map out the various aspects of sexism as seen in the everyday social constructs and use it to comprehensively map them out for the benefit of the identification tasks \citep{samory_sen_2021}.

Also looking beyond the text, \citet{chiril_etal_2020} performed further characterization of the binary sexist classification by distinguishing cases where the addressee is directly addressed from those where she is not. The three categories being: (i) directed assertions - sexist tweet directly addressed to a woman or a group of women; (ii) descriptive assertions - sexist tweets not directed to an addressee; and (iii) reported assertions - tweets containing report of an experience or a denunciation of sexist behavior. On performing classification based on results per class, they identified the absence of context with the utterance, humor, and satire, and the use of stereotypes or metaphors to be the causes of misclassification through their manual error analysis in their best performing model- BERT. As \citet{frenda_2019} had also stated one of their principle problems is the use of linguistic devices like irony and sarcasm. In general, it has been highlighted in multiple research (e.g., \citep{guest-etal-2021, rodriguez_2020}) that the more subtle forms of sexism (mostly depended on the context) are not picked up well by the models. As \citet{singh-etal-2021} note in their error analysis, many of the confounding variables were specific lexical items that either denoted explicitly sexual content or conveyed strongly negative sentiment.

Different psychometric scales can also be used to map out various aspects of sexism/misogyny as a social construct, to comprehensively detect the different categorizations. \citet{king_king_1997} reaffirm the previously stated theory on modern sexists, and describe them as “people who while rejecting old-fashioned discrimination and stereotypes, may believe that discrimination against women is a thing of the past, feel agnostic against women who are making political and economic demands, and feel resentment about special favors for women, such as policies designed to help women in academics and work.” In other words, the distinction between old-fashioned and modern sexism lies in the fact that the former showcases an obvious unequal treatment of women while questioning their intelligence, while the latter is less sympathetic to women’s issues (if at all they perceive them to be issues) since they presume greater equality in the workforce than what exists. The Modern Sexism (MS) scale this study provides aims to be a good indicator to detect modern sexism, which could be both overt and covert in nature. People endorsing MS beliefs are hence less likely to detect the occurrence of a normative sexist behavior \citep{swim_mallett_2004}. In the review by \citet{swim_cohen_1997} on the MS scale, they indicate the same as they observe that it measures the subtle forms of sexism that are built upon cultural and societal norms. They also review another general measure of sexism, namely the Attitude Toward Women Scale (AWS), which measures overt or blatant sexism. And through their analysis, they indicate that even with these distinctive differences, both share related constructs. These social constructs are often perpetrated as discriminatory attitudes towards a feminine gender role, which are traditionally allocated and differentiated by sex. \citet{garcia_etal_2015} propose a scale to assess the gender role attitude, showing how sexist attitudes can be modified using the theoretical perspectives of gender equality.

\subsubsection{Linking social science theories to computer science research}
Following from the previous subsection where we introduced our argument that sexism/misogyny is not a binary task, in this section, we expand on that point by providing social science theories and scales to explain the need to not computationally limit the classification to the binary output. To support that, alongside including the theories and scales, we also analyze how some studies have aided their work with these theories in any capacities (i.e., the extent of adaptation - using one or more categories of the scaling) and implemented them at any stage of their research. We distinguished each subdivision into two parts: the \textbf{concept} and the \textbf{applications}, to help us differentiate between the concepts themselves and on how they are implemented in studies. 

\paragraph*{Sexism is not always hostile}

\noindent\textbf{Concept.}\\
\citet{grosz_conde_2020} state, that models can perform detection tasks easier on datasets containing large amounts of ``hostile” sexism, since it hinges on some words, regardless of their context. But that does not provide a real-world scenario. In general, sexism is said to have two components: hostility towards women and endorsement of traditional gender roles, and most of the sexist attitude measures so far have stemmed from there. But it is not always so. Through their anthropological research on sexism, \citet{glick_fiske_1997} call sexism ``fundamentally ambivalent”, adding the subjectively benevolent nature of sexism to the previously perceived singularly hostile nature. They argue that the ``simultaneous existence of male structural power and female dyadic power” creates an ambivalent ideology. While the hostile ideology seeks justification of their male position through derogatory characterization of women (HS), benevolent ideology relies on kinder and gentler justification, which may inherently look as subjectively positive for the sexist as they encompass feelings of protectiveness and affection towards women (BS). By drawing parallels from paternalism, which also has two ideologies- dominative and protective, they demonstrate that the protectiveness is particularly strong when women(e.g., wives, mothers, daughters) are dyadically dependent on men, as a feeling something akin to the sense of ``ownership”. The hierarchical stereotype ideology explained before constitutes the belief contributing to the gender differentiation. Like paternalism, it also consists of both hostile and benevolent side. Competitive gender differentiation being the hostile kind, delves on negative stereotypes of women implying men to be the better gender; and the complementary gender differentiation (the benevolent kind) stems from the traditional stereotypes of women through assigned gender roles and men’s dyadic dependence on women, albeit in an extremely positive light \citep{eagly_mladinic_1994}. Similarly, for heterosexuality, which has a hostile side when viewing women as mere sexual objects who use sexual attraction to gain power over men; and intimate or benevolent side that romanticizes the former belief, viewing women as necessary for men to feel “complete”. \\

\noindent\textbf{Applications.}\\
Sexism in ambivalent theory \citep{glick_fiske_1996} is thus hypnotized to encompass these three sources of male ambivalence, which has been used by \citet{jha-mamidi-2017} to computationally identify benevolent sexism, and classify sexist content based on the two components. They confirm the hypothesis that HS is evidently negative and easily identifiable, while BS is retweeted much more and is camouflaged, seemingly harmless or noble and hence, harder to detect. It was seen that while SVM showed high precision for both, recall was quite low for HS; their Seq2Seq model (LSTM-based bi-directional RNN) showed a higher recall for both, even though its precision was not as high, presumably because it takes in the structure of the tweet. But owing to the bag-of-n-grams feature of FastText (and lesser parameters to tune), it outperformed both the former classifiers. On the other hand, \citet{singh-etal-2021} used the hostile side of the three sources of male ambivalence to define sexism binarily and annotate dialogues in popular sitcoms. Using these concepts, they manually annotated the external datasets (source domain) and used a semi-supervised domain-adaptive learning approach to generate classes in the model for the unannotated data (target domain), thus further augmenting the training data and improving the final classification performance. However, error analysis showed certain false positives like incorrectly classifying aggressive negative statements to a particular woman, or contents with explicit sexual terms and mentions of marriages or weddings as sexist. This could be the underlying drawback of not using a diverse dataset since the authors had included dialogues that included derogatory terms and dialogues justifying stereotypes against women or gender roles. But \citet{mishra_2019} use the concepts from previous research rather differently, by taking inspiration from studies that use randomly initialized user embeddings for improving performances, and inter and intra-user representations based on tweets. Instead of the former semi-supervised approach, they use graph convolutional networks (GCN) based approach, applied to the heterogenous graph representation of two types of nodes- authors and their tweets, to generate richer author profiles. The intention was to use such heterogenous representation to enable the model to learn both community structure and linguistic behavior of authors in such communities. Even with this improvement, several abusive tweets were misclassified, primarily due to the presence of abusive content in the URL (not in the tweet itself), and the deliberate obfuscation of words and phrases by the authors to evade detection.

\paragraph*{Subtle forms of sexism/misogyny}

\noindent\textbf{Concept.}\\
Since most of the sexism measurement scales are focused on hostile sentiments, it fail to capture the contemporary forms of subtle sexism, which are often cloaked in the guise of egalitarian views and harbor (more) traditional beliefs. Only some of the previous works (such as \citep{rodriguez_2021, rodriguez_2022, samory_sen_2021}) have addressed the more subtle and covert forms of sexism. Yet, due to the increase in social awareness of sex discrimination, the more blatant form of sexism is reduced, replaced with the subtle forms of indirect indices. And the lack of conceptual framework of understanding, coupled with methodological problems were indicated in the simulation study conducted by \citet{beattie_diehl_1979}, where they observe the use of indirect means to interpret the gender and hence influence the evaluation criteria. This gave suggestive evidence to a new form of sexism called ``neosexism'', which was first introduced by \citet{tougas_brown_1995}, and defined as ``a manifestation of a conflict between egalitarian values and residual negative feelings towards women''. They used a predictive model of ‘attitude to affirmative actions’ to test the discriminatory bias and evaluated the practical implications of neosexism through their Neosexism Scale (NS). The study indicated that ``neosexist beliefs were linked with opposition to programs designed to facilitate integration of both women and minorities'', which leads to further proves the importance of understanding the existing prejudicial beliefs of women to understand the different forms of sexism. \\

\noindent\textbf{Applications.}\\
An analysis of the cross-sectional data during the 2016 US presidential election and the \#MeToo movement by \citet{archer_kam_2020} shows its significant correlation to neosexism, and the various degrees of dismissal of the respondents to the existing gender discrimination, hence indicating its existence in online platforms. \citet{zeinert-etal-2021} had used NS in their work on Danish tweets to add neosexism to their taxonomy along with the previously categorized forms of sexism. Interestingly, while annotating, they found that neosexism formed the most common form of misogyny and accounted for most of the annotation challenges based on disagreements, primarily due to the challenge of understanding the author’s intentions, the degree of abuse (since misrepresentation could harm the subject or the fact) and lack of world knowledge. This further added to the class imbalance in the last stage of sexism labeling in their dataset which affected the reliability of the performance, even though they started with a 1:1 class balance at the initial stage (labeling abusive or not) of their iterative labeling scheme based on the MALER framework proposed by \citet{finlayson_2017}. To prevent such bias caused by an imbalanced dataset, \citet{indurthi_etal_2019} process the training dataset using SMOTE \citep{chawla_2002} which synthetically oversamples data and ensure all classes have an equal number of instances. While the existence of subtle forms of sexism and misogyny is undeniable, having unbiased data representative of the same is essential to gain a better computational outcome.


 


\section{Summary of general strategies used and existing challenges}

Our summary of key research findings identified through the literature review reflects the current drawback in the study of sexism and misogyny identification tasks. Irrespective of the different measures taken by the literary works, some limitations remain consistent, which further hinder obtaining a robust model capable of quantifying sexism or misogyny. As \citet{vidgen_2020} suggest, ``More standardization is an important aspiration as research continues to mature, although it must be balanced with enabling research innovation and freedom.'' Therefore, we summarise the research findings in the following points:
(1) We identify the various forms of online sexism -- from direct abuse to implicit biases, indicating a need for nuanced detection and classification mechanisms; (2) Most studies focus on Western contexts and often overlook intersectional identities such as race, age, and sexual orientation, limiting the generalizability of findings; (3) There is inconsistency in how online sexism is defined and studied, with diverse methodologies and terminology making comparative analysis difficult; (4) The complex nature of online sexism requires insights from multiple disciplines -- such as Computer Science and Social Science to form holistic solutions; (5) While some automated solutions for such a huge amount of online content show promise, there is a need for theoretically grounded exploration of online sexism beyond the scope of current research; (6) A high performance score does not necessarily indicate a robust model, which consequently highlight the potential need to incorporate additional metrics for a more comprehensive evaluation.

\section{Conclusion}\label{sec:conclusion}
In this systematic literature review, we examined the multifaceted and evolving phenomenon of online sexism and misogyny, with particular attention to its categorization, detection, and the methodologies employed by existing literature across diverse digital platforms. Our analysis revealed that online sexist discourse manifests in a wide spectrum, ranging from explicit abuse and harassment, to implicit and context-dependent expressions. While advancements in NLP and machine learning have enabled meaningful progress in detecting overt forms of online sexism, subtler and more nuanced forms remain underexplored to maintain good consistency and accuracy using various methodologies.
We also found that existing taxonomies and datasets are often limited in scope, predominantly rooted in Western contexts (both culturally and linguistically), and frequently lack intersectional perspectives. Furthermore, we observed that the research in this domain remains largely siloed, with insufficient integration between computational, sociological, and feminist theoretical frameworks.
The findings of this review emphasizes the necessity for more inclusive, interdisciplinary, and context-aware approaches to the study and mitigation of online sexism. Future research should prioritize the creation of culturally diverse datasets, the development of unified conceptual taxonomies taking into account the range of expressions for sexism and misogyny, and possible incorporation of marginalized voices in both the design and evaluation of detection systems for better perceptivity. By fostering collaborations across disciplines and centering equity in technological development, we can advance more effective and socially responsive strategies to combat sexism and misogyny in online digital spaces. By highlighting these challenges, we aim to guide future research toward deeper theoretical integration, improved computational methods, and more representative datasets, ultimately fostering more nuanced understandings of sexism and misogyny in digital spaces.
Through this work, we hope to contribute to further development on this topic ensuring updated resources on the same, and encouraging investigation on the change in dynamics of online sexism and misogyny.

\section*{Declarations}


\paragraph*{Material and code availability}
\noindent No data was generated during this research, but were acquired from online websites or through the API access of the stated citation databases. 
All the shareable acquired data from the databases collected and used in the research, along with its analysis is made available in the GitHub page: \url{https://github.com/booktrackerGirlsys-lit-review-Sexism}\footnote{This repository would be made public upon acceptance for publication.}. We also include the permissive license to allow users to use, modify and distribute the materials. 

\paragraph*{Competing interests}
\noindent The authors have no competing interests to declare that are relevant to the content of this article.

\paragraph*{Funding}
\noindent We thank the University of Exeter for funding the cost to access the Web of Science Expanded API and SerpAPI.
A.D.'s time on the research was funded by the SSIS Global Excellence PhD Studentship from the University of Exeter.
S.B.'s time on the research was funded by the European Research Council (ERC) under the European Union’s 
Horizon 2020 research and innovation programme (grant agreement No 101019284).
C.Q.C. thanks the Ewha Frontier 10-10 project and the DSO National Laboratories Singapore for funding this research.


\bibliography{references}
\bibliographystyle{apalike}

\pagebreak
\appendix
\setcounter{equation}{0}
\setcounter{figure}{0}
\setcounter{table}{0}
\setcounter{page}{1}
\makeatletter
\renewcommand{\bibnumfmt}[1]{[S#1]}
\renewcommand{\citenumfont}[1]{S#1}

\section{Systematic Literature Review strategy}
\subsection{Draft search string}\label{draft_search_string}

\textit{Draft string length: 256 character limit}
\begin{enumerate}
    \item (misogyny OR sexism) 
    \item (hate OR toxic OR abusive OR offensive)
    \item (detection OR identification OR prediction OR classification) 
    \item ("natural language processing" OR NLP OR "deep learning" OR "machine learning" OR ML OR "artificial intelligence" OR AI)
    \item 1/ AND 2/ AND 3/ AND 4/
    \item Limit 5 to (english language and yr="2012 -Current")
\end{enumerate}

\subsection{Inclusion  and exclusion criteria}\label{exclusion_inclusion_criteria}
\begin{enumerate}
    \item Remove posts from online publishing platforms, online research platforms or similar (e.g. blogs)
    \item Remove papers outside the year range (2012-2022)
    \item Remove papers not written in English
    \item Remove dissertations, theses, books, and whole conference proceedings; but include pre-prints within the period
    \item Remove symposium submissions
    \item Limit by date of external events (2000- current)
    \item Limit by the platform used for study- comparative study across platforms maybe included
    \item Remove studies not looking at text data (so images, video, etc)
    \item Remove studies that look into offline instances of sexism and misogyny
    \item Remove studies that do not look into online social platforms (like Meta, Twitter, Reddit, etc.)
    \item Remove studies that focus on the mental and physical impact of online hate speech from the aforementioned platforms.
    \item Only keep papers that measure misogyny and/or sexism.
    \begin{enumerate}
        \item This means removing studies with no quantitative methods, papers proposing guidelines, policy recommendations, discussions, tutorials, dataset descriptions, research briefs, working papers, purely theoretical approaches, opinion pieces, position papers, case studies, etc.
        \item Removing studies where frameworks are only stated without any measurements/results proceeding it.
        \item It will include papers that measure misogyny/sexism with other forms of online hate, such as toxicity, hate-speech, aggression, etc.
        \item It can include gender-bias classification studies that fall close to the definition of sexism/misogyny as generic terms, depending on the context it is being used.
    \end{enumerate}

\end{enumerate}

\section{Citation Database queries}
\begin{table}[h!]
    \centering
    \begin{tabular}{ |p{2cm}|p{13cm}|  }
        \hline
        \multicolumn{2}{|c|}{\textbf{Citations and their search queries}} \\ 
        \hline\hline
        \begin{center}{\textbf{Google Scholar}}\end{center} & \footnotesize ((misogyny OR sexism) AND (hate OR toxic OR abusive OR offensive) AND (detection OR identification OR prediction OR classification) AND (“natural language processing" OR NLP OR "deep learning" OR "machine learning" OR ML OR "artificial intelligence" OR AI) AND (language="English" AND yr="2012 -2022")) \\ 
        \hline
        \begin{center}{\textbf{ArXiv}}\end{center} & \footnotesize ($all:sexism+OR+all:sexist+OR+all:misogyny+OR+all:misogynist+OR+all:\%22gender+discrimination\%22+OR+all:\%22gender+violence\%22+OR+all:\%22gender+stereotype\%22$)\\ 
        \hline
        \begin{center}{\textbf{Elsevier}}\end{center} & \footnotesize ('misogyny detection OR misogyny identification OR misogyny prediction OR misogyny classification OR sexism detection OR sexism identification OR sexism prediction OR sexism classification') \\ 
        \hline
        \begin{center}
            {\textbf{Scopus}}
        \end{center} & \footnotesize TITLE-ABS-KEY (( misogyny OR sexism OR gender AND violence OR gender AND discrimination ) AND ( detection OR identification OR prediction OR classification ) AND PUBYEAR $>$ 2011 AND PUBYEAR $<$ 2023 AND PUBYEAR $>$ 2011 AND PUBYEAR $<$ 2023 AND ( LIMIT-TO ( SUBJAREA , "SOCI" ) OR LIMIT-TO ( SUBJAREA , "COMP" ) OR LIMIT-TO ( SUBJAREA , "PSYC" ) ) AND ( LIMIT-TO ( LANGUAGE , "English" )) \\
        \hline
        \begin{center}{\textbf{Semantic Scholar}}\end{center} & \begin{center}(‘online sexism misogyny’)\end{center} \\ 
        \hline
        \begin{center}{\textbf{Web of Science}} \\ (Social Science)\end{center} & \footnotesize TS=((misogyn* OR sexis* OR (gender NEAR/10 discrim*) OR (gender NEAR/10 stereoty*) OR (gender NEAR/10 violence) OR (gender NEAR/10 based)) NEAR/200 (detect* OR identif* OR predict* OR classif*)) AND WC=((“History” OR “Political Science” OR “Women\'s Studies” OR “Social Sciences” OR “International Relations” OR “History \%26 Philosophy Of Science” OR “Linguistics” OR “Anthropology” OR “Sociology” OR “Social Work” OR “Language \%26 Linguistics” OR “Information Science” OR “Psychology” OR “Social” OR “Ethnic Studies” OR “Philosophy” OR “Psychiatry”) NOT (“Computer Science” OR “Artificial Intelligence” OR “Theory \%26 Methods” OR “Engineering” OR “Software Engineering” OR “Scientific Disciplines” OR “Automation \%26 Control Systems” OR “Mathematical” OR “Mathematics” OR “Mathematical Methods”)) AND PY=2012-2022\\ 
        \hline
        \begin{center}{\textbf{Web of Science}} \\ (Computer Science)\end{center} & \footnotesize TS=((misogyn* OR sexis* OR (gender NEAR/10 discrim*) OR (gender NEAR/10 stereoty*) OR (gender NEAR/10 violence) OR (gender NEAR/10 based)) NEAR/200 (detect* OR identif* OR predict* OR classif*)) AND WC=((“Computer Science” OR “Artificial Intelligence” OR “Theory \%26 Methods” OR “Engineering” OR “Software Engineering” OR “Scientific Disciplines” OR “Automation \%26 Control Systems” OR “Mathematical” OR “Mathematics” OR “Mathematical Methods”) NOT (“History” OR “Political Science” OR “Women\'s Studies” OR “Social Sciences” OR “International Relations” OR “History \%26 Philosophy Of Science” OR “Linguistics” OR “Anthropology” OR “Sociology” OR “Social Work” OR “Language \%26 Linguistics” OR “Information Science” OR “Psychology” OR “Social” OR “Ethnic Studies” OR “Philosophy” OR “Psychiatry”)) AND PY=2012-2022\\
        \hline
    \end{tabular}
    \caption{\centering Citation databases and their respective queries}\label{table:citations}
\end{table}

\section{Terminologies and their meaning}
\begin{table}[htbp]
    \centering
    \footnotesize
    \begin{tabular}{|p{3cm}|p{2cm}|p{9cm}|}
    \hline
    \Centering Construct & \Centering - & ``A construct is an abstract concept that is specifically chosen (or `created') to explain a given phenomenon. Constructs used for scientific research must have precise and clear definitions that others can use to understand exactly what it means and what it does not mean.'' \citep{bhattacherjee2019social}\\
    \hline
    \Centering Computational Social Science & \Centering CSS & Computational social science is an interdisciplinary academic sub-field concerned with computational approaches to the social sciences. It leverages the capacity to collect and analyze data with an unprecedented breadth and depth and scale. \citep{lazer_2020}.\\
    \hline
    \Centering Hostile Sexism & \Centering HS & \Centering Hostile sexism refers to negative views toward individuals who violate traditional gender roles. For example, some people disparage girls who enter traditionally masculine domains such as science or sports \citep{daniels_etal_2011}. \hspace{\textwidth} Part of ambivalent sexism \citep{glick_fiske_1996}. \\
    \hline
    \Centering Neo-sexism Scale & \Centering NS & \Centering A scale designed to tap into a new type of gender prejudice, called neo-sexist beliefs \citep{tougas_brown_1995}.\\
    \hline
    \Centering Benevolent Sexism & \Centering BS & \Centering  Benevolent sexism includes valuing feminine-stereotyped attributes in females (e.g., nurturance) and a belief that traditional gender roles are necessary to complement one another. Benevolent sexism also includes the view known as paternalism that females need to be protected by males. Benevolent sexism contributes to gender inequality by limiting women's roles \citep{daniels_etal_2011}. \hspace{\textwidth} Part of ambivalent sexism \citep{glick_fiske_1996}.\\
    \hline
    \Centering Bidirectional Encoder Representations from Transformers  & \Centering BERT & \Centering BERT is a language representation model, which is designed to pre-train deep bidirectional representations from unlabeled text by jointly conditioning on both left and right context in all layers \citep{devlin_2018}.\\
    \hline
    \Centering Language Models (or Large Language models) & \Centering LM (or LLM) & \Centering A large language model is a computational model capable of language generation or other natural language processing tasks. As language models, LLMs acquire these abilities by learning statistical relationships from vast amounts of text during a self-supervised and semi-supervised training process.\\
    \hline
    \Centering Bag-of-words & \Centering BoW & \Centering ``A bag-of-words is a representation of text that describes the occurrence of words within a document. It involves two things: 1. A vocabulary of known words. 2. A measure of the presence of known words."\url{http://tinyurl.com/5n6d9knt} \\
    \hline
    \Centering Descriptive Paradigm & \Centering - & \Centering  ``The descriptive paradigm encourages annotator subjectivity to create datasets as granular surveys of individual beliefs. Descriptive data annotation thus allows for the capturing and modeling of different beliefs.'' \citep{röttger_2022}\\
    \hline
    \Centering Perspective Paradigm & \Centering - & \Centering ``The prescriptive paradigm, on the other hand, discourages annotator subjectivity and instead tasks annotators with encoding one specific belief, formulated in the annotation guidelines. Prescriptive data annotation thus enables the training of models that seek to consistently apply one belief.'' \citep{röttger_2022} \\
    \hline
    \end{tabular}
    \caption{Terminologies}\label{table:terminologies}
\end{table}

\section{Experimentation results from other citation databases}\label{sup_sec:other_cits}

For \textbf{Google Scholar}, we used both external APIs like SerpAPI for scraping the data, as well as a software named 'Publish or Perish' \citep{publish_or_perish} to collect the search results. Both of the methods were rejected because of their disadvantages. Such as, Publish or Perish could only extract 1000 results at a time for each search query. While this drawback was overcome by searching for documents with a shorter range of years to stay within the limit, it lacked some of the fields that were needed for this study - abstract and discipline. Alternatively, SerpAPI \citep{SerpAPI_2019} worked similar to a web scrapping tool and could only scrape the results as the search engine demonstrates, i.e., it only scrapes what Google shows on their Google Scholar pages, nothing more. Even though the fields we got through this API were relevant, they did not contain the full information we needed for the analysis. For example, the full text in the title and abstract was missing and was instead indicated with dotted extensions in the beginning and end of the text.

\section{Web of Science strategy}

\noindent We performed automated elimination (or pre-processing) techniques based on the following criteria to narrow down our search results for both areas of study\footnote{More details can be found here: \url{https://images.webofknowledge.com/images/help/WOS/hp_advanced_search.html}}: \label{sup:wos_search}
\begin{itemize}
    \item Remove studies that are not published in English.
    \item Remove studies that do not contain any abstracts.
    \item Keep only the first abstract in studies that contain more than one abstract.
    \item Remove certain publication types, such as review articles and editorials.
\end{itemize}

\noindent With the Web of Science API, separate search queries were used for the two broad disciplines ( or research areas) - CS and Social Science. The categories of the research areas taken for each of them are as follows:
\\

\begin{table}[htbp]
    \centering
    \begin{tabular}{ |p{6cm}||p{6.5cm}|  }
        \hline
        \Centering \textbf{\Large Computer Science} & \Centering \textbf{\Large Social Science}\\
        \hline
        \begin{itemize}
            \item Computer Science
            \item Artificial Intelligence
            \item Theory and Methods
            \item Engineering
            \item Software Engineering
            \item Scientific Disciplines
            \item Automation and Control Systems
            \item Mathematical
            \item Mathematics
            \item Mathematical methods
        \end{itemize} & 
        \begin{itemize}
            \item History 
            \item Political Science 
            \item Women\'s Studies 
            \item Social Sciences 
            \item International Relations
            \item History and Philosophy of Science 
            \item Linguistics
            \item Anthropology
            \item Sociology
            \item Social Work
            \item Language and Linguistics
            \item Information Science
            \item Psychology
            \item Social
            \item Ethnic Studies
            \item Philosophy
            \item Psychiatry
        \end{itemize} \\ 
        \hline
    \end{tabular}
    \caption{Categories for each area of research}
\end{table}\label{table:category_disc} 

\noindent These disciplines were taken from the Web of Science category list, which branches from five major research areas - out of which we took the two categories \textbf{Social Sciences} and \textbf{Technology}. The published works present in the Web of Science Core Collection are assigned to at least one Web of Science category. Each of the said Web of Science categories (as listed in table \ref{table:category_disc}) is mapped to one research area found in the classification of research areas\footnote{Source: \url{https://images.webofknowledge.com/images/help/WOS/hp_research_areas_easca.html}}.

\section{ArXiv strategy}
\noindent The ArXiv API was used following the query search strategy\footnote{More details of the search strategy can be found here: \url{https://info.arxiv.org/help/api/user-manual.html\#query_details}}. \label{sup:arxiv_search}

\noindent We performed automated elimination (or pre-processing) techniques based on the following criteria to narrow down our search results for both areas of study:
\begin{itemize}
    \item Remove studies that are not between 2012 and 2022.
    \item Remove studies that do not contain any abstracts.
\end{itemize}

\noindent While combining search results of \ref{sup:wos_search} and \ref{sup:arxiv_search}, care was taken to remove the duplicate studies based on the title and abstract, where we kept the study from the former database. This is to ensure consistency along the data since the published and updated (i.e., when the pre-prints were submitted to ArXiv) years could differ, hence ensuring the published works are not mislabeled as pre-prints.

\section{Further analysis of the initial search results}

\subsection*{Documents by disciplines}

\begin{figure}[h!]
  \centering
    \begin{minipage}[b]{0.5\textwidth}
        \includegraphics[width=\textwidth,height=5cm]{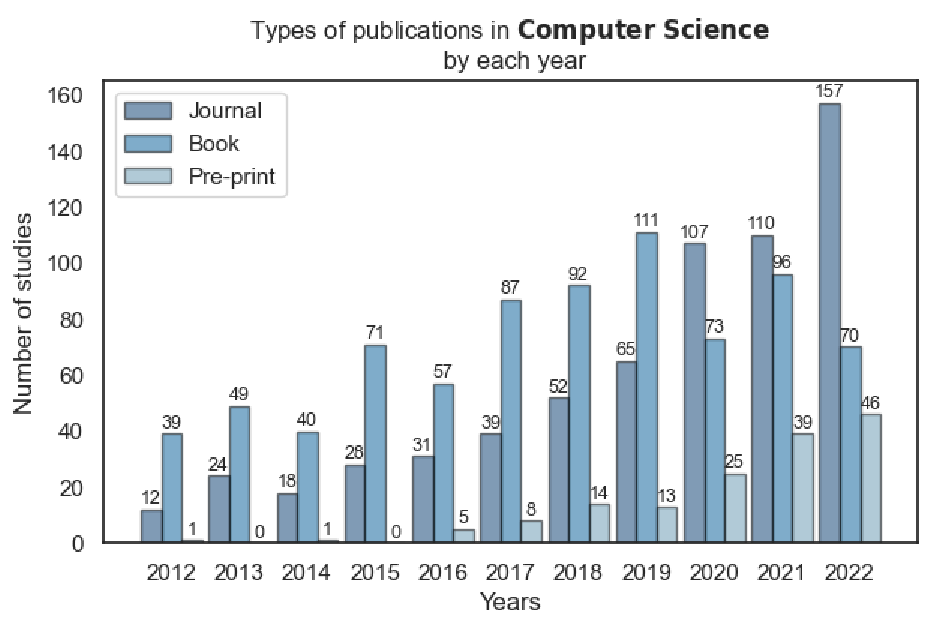}
        \caption*{\centering (a)}
        \label{img:pub_type_comp}
      \end{minipage}
      \hspace{-10pt}
      \hfill
      \begin{minipage}[b]{0.5\textwidth}
        \includegraphics[width=\textwidth,height=5cm]{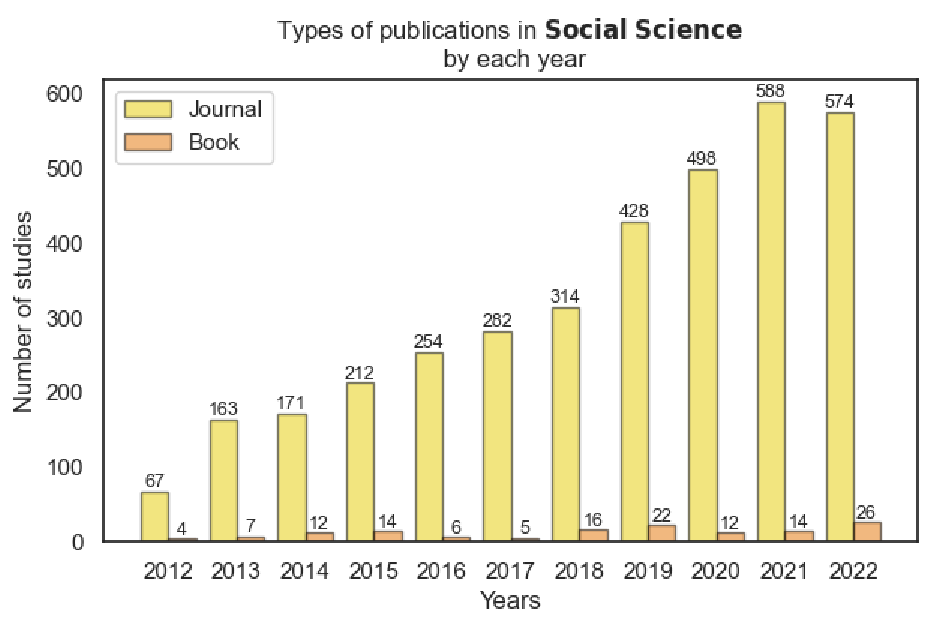}
        \caption*{\centering (b)}
        \label{img:pub_type_soc}
    \end{minipage}
    \hfill
      \caption{\centering (a) Type of publications in Computer Science. \hspace{\textwidth}(b) Type of publications in Social Science.}
      \label{fig:pub_types_disc}
\end{figure}

Figure \ref{fig:pub_types_disc} shows the frequency of publications per year in the range of 2012-2022, as per each discipline and publication type. Like we had discussed previously in Section \ref{sec:data_overview}, we see a huge disparity in the number of publications between the disciplines which focus on sexism and/or misogyny. This inherently appears to impact on the diversity of the concept explored by the disciplines, with SS exploring a broader range of themes than CS. Furthermore, we also see that the type of publication too differs quite a bit as CS tend to produce a handful of research as pre-prints on this topic.

\hfill\break

\subsection*{Documents focused on social media platforms}

\begin{figure}[h!]
  \centering
    \begin{minipage}[b]{0.5\textwidth}
        \includegraphics[width=\textwidth,height=4cm]{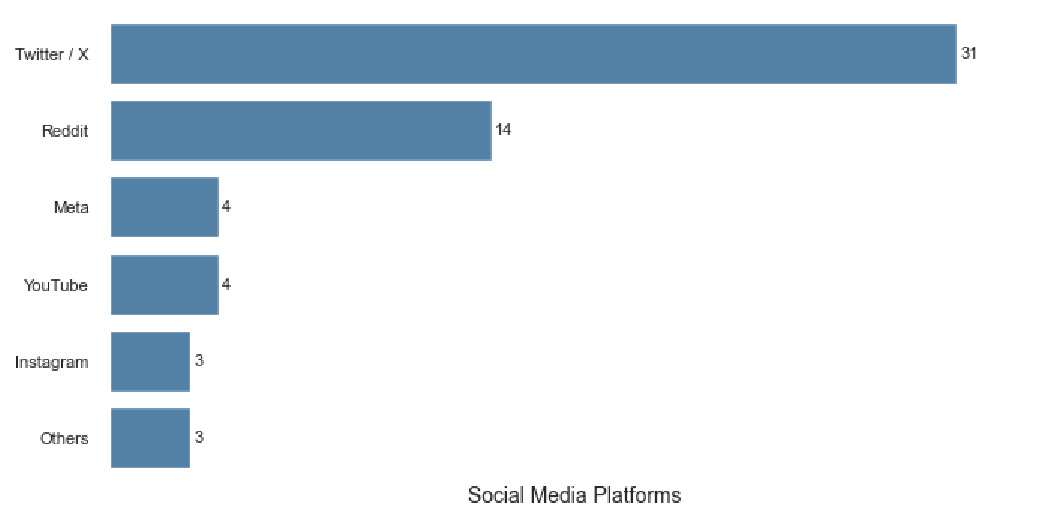}
        \caption*{\centering (a)}
        \label{img:soc_media_comp}
      \end{minipage}
      \hspace{-10pt}
      \hfill
      \begin{minipage}[b]{0.5\textwidth}
        \includegraphics[width=\textwidth,height=4cm]{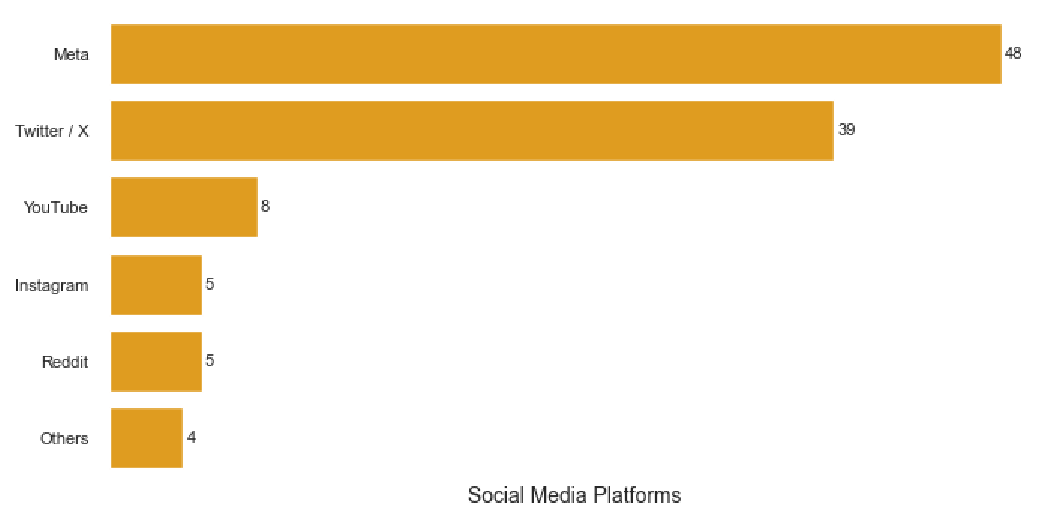}
        \caption*{\centering (b)}
        \label{img:soc_media_soc}
    \end{minipage}
    \hfill
      \caption{\centering Publications mentioning social media platforms in titles or(/and) abstracts in: \hspace{\textwidth}(a) Computer Science. \hspace{\textwidth}(b) Social Science.}
      \label{imgs:pub_types}
\end{figure}
The share of documents focusing on different social media platforms, as observed in Figure \ref{imgs:pub_types}, reveal that X (formerly Twitter) was the dominant platform for most research in CS, while Facebook (or Meta) was more dominant in SS till 2022. The ease of access to Twitter data during the period could have been a contributing factor to allow application of automated approaches in CS. Whereas, Facebook having more number of active users could have contributed to more research in SS, than any other platforms (including Twitter).

\hfill\break

\subsection*{General topics centering around sexism or misogyny}

\begin{figure}[h!]
  \centering
    \begin{minipage}[b]{0.5\textwidth}
        \includegraphics[width=\textwidth,height=5cm]{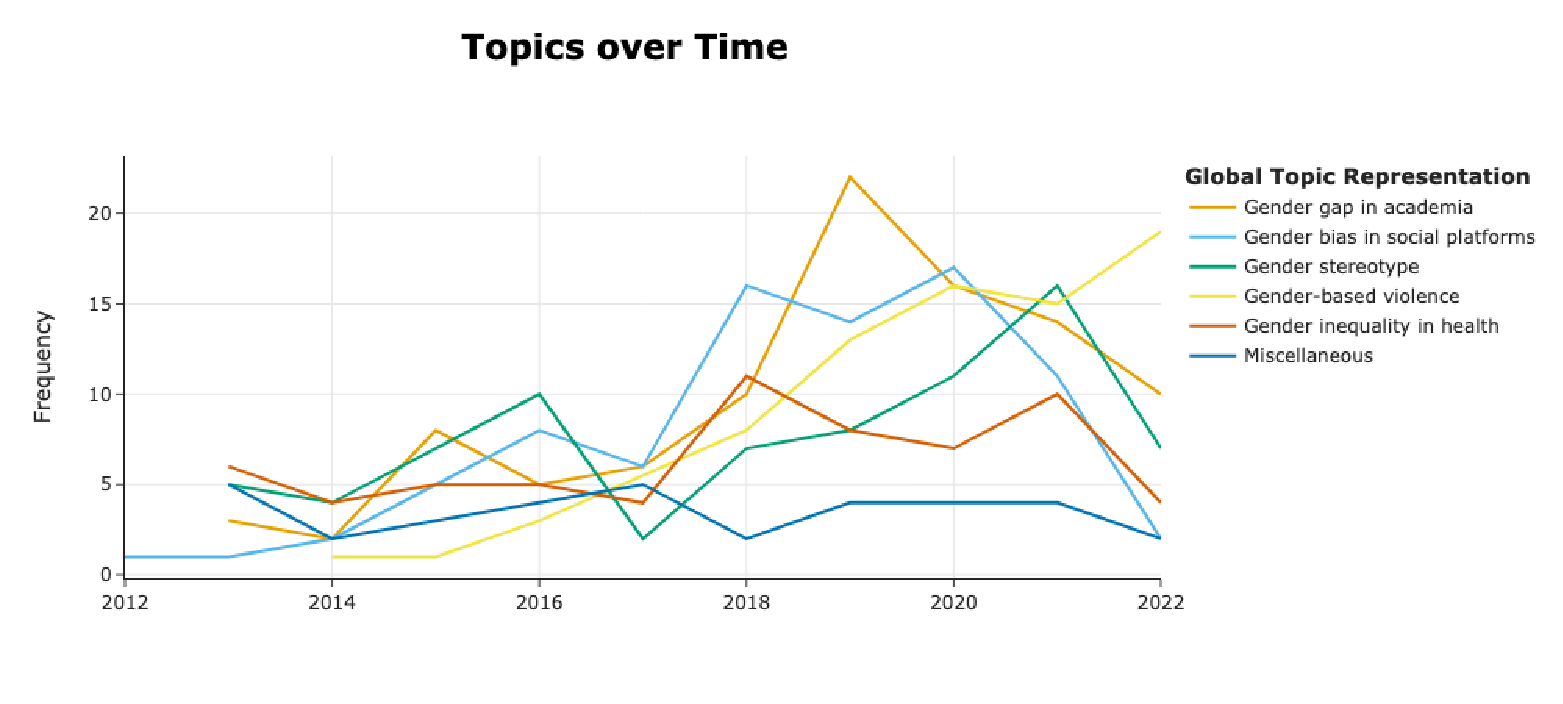}
        \caption*{\centering (a)}
      \end{minipage}
      \hspace{-10pt}
      \hfill
      \begin{minipage}[b]{0.5\textwidth}
        \includegraphics[width=\textwidth,height=5cm]{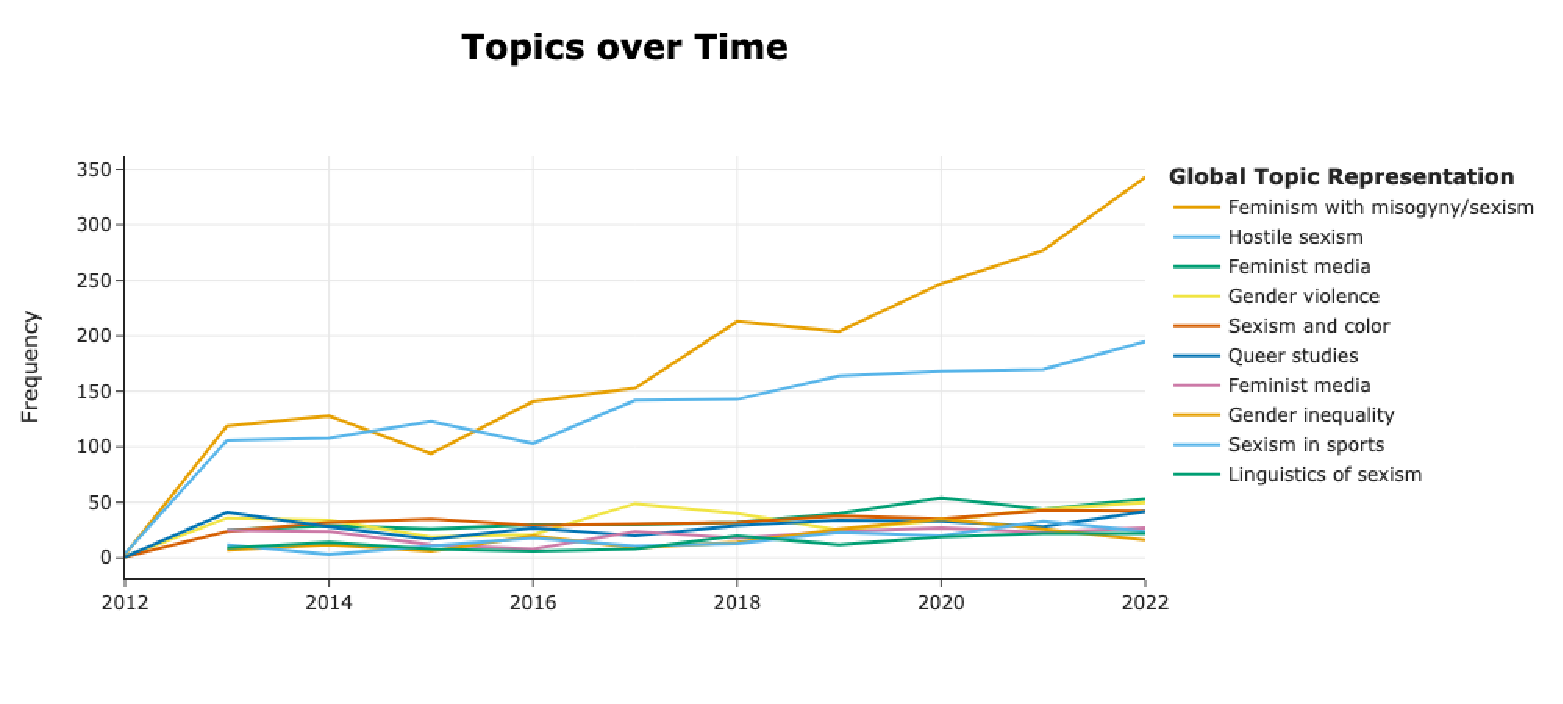}
        \caption*{\centering (b)}
    \end{minipage}
    \hfill
      \caption{\centering General topics centered around sexism/misogyny over the years in: \hspace{\textwidth}(a) Computer Science. \hspace{\textwidth}(b) Social Science.}
      \label{imgs:theme_types}
\end{figure}

Figure \ref{imgs:theme_types} show the different thematic (or topic) representations across the disciplines over the period of 2012-2022. Not only do we see a wider range of themes in SS expanding over more number of research (like we observe in the previous subsection as well), but also a steady rise in most of the topics along the time. Especially the theme of `Feminism with misogyny/sexism' and the `Hostile sexism' seems to be of particular interest for SS research, given the proliferation of sexism and misogyny beyond offline spaces. The theme
of `Linguistics in sexism' are instrumental in capturing the subtle forms of sexism, and is therefore seen to gain traction over the years. The themes of CS research on sexism and misogyny seem to fluctuate in the given period with no consistent rise, except for the `Gender-based violence'. 

\section{Abbreviation of models}\label{abbr:models}
The abbreviations used in Figure \ref{fig:sankey} are a collection of the following models as shown in Table \ref{tab:abbr_mod}.
\begin{table}[h!]
\setlength{\tabcolsep}{1pt}
    \centering
    \footnotesize
    \begin{tabular}{ |p{2cm}|p{10cm}|  }
    \hline
    \centering Abbreviation & Full name of the model(s) \\
    \hline
    \centering LR & Logistic Regression \\
    \centering RF & Random Forest \\
    \centering SVM & Support Vector Machine\\
    \centering BERT & BERT, RoBERTa, mtBERT, FlauBERT, XLMRoBERTa, BERT-base, among other BERT based models \\
    \centering CNN & Convolutional neural network \\
    \centering NB & Naïve-Bayes, MultinomialNB \\
    \centering LSTM & LSTM, Bi-LSTM \\
    \centering W2V & Word2Vec, GloVe \\
    \centering LDA & Latent Dirichlet Allocation \\
    \centering GB & Gradient Boosting, CatBoost \\
    \centering DT & Other Decision Tree models \\
    \centering GCN & Graph Convolutional Network \\
    \centering RNN & Recurrent Neural Network\\
    \centering DNN & Deep neural network (unspecified)\\
    \centering XGB & XGBoost \\
    \centering kNN & k-NearestNeighbours \\
    \centering BoW & Bag-of-Words \\
    \centering RC & Ridge Classifier \\
    \centering n-grams & unigrams, bi-grams and other types of n-grams \\
    \centering IG & Information Gain \\
    \centering MLP & Multi-layer Perceptron \\
    \centering Embeddings & FastText, InferSent, Universal Sentence Encoder, and other types of embeddings \\
    \centering OVR & One-vs-Rest \\
    \centering GRU & Gated Recurrent Units \\
    \hline
    \end{tabular}
    \caption{Model names and their abbreviation}
    \label{tab:abbr_mod}
\end{table}

\section{Most frequent keywords}\label{sec:most_freq_key}
In this section, we demonstrate the top 100 keywords in the co-occurrence network, like in Section \ref{subsec:keywrd_co} but based on all the manuscripts of each individual field.

\subsection{Most frequent keywords in Computer Science}\label{sub_sec:keywords_comp}
\begin{figure}[htbp]
    \centering   
    \includegraphics[width=0.9\textwidth, keepaspectratio]{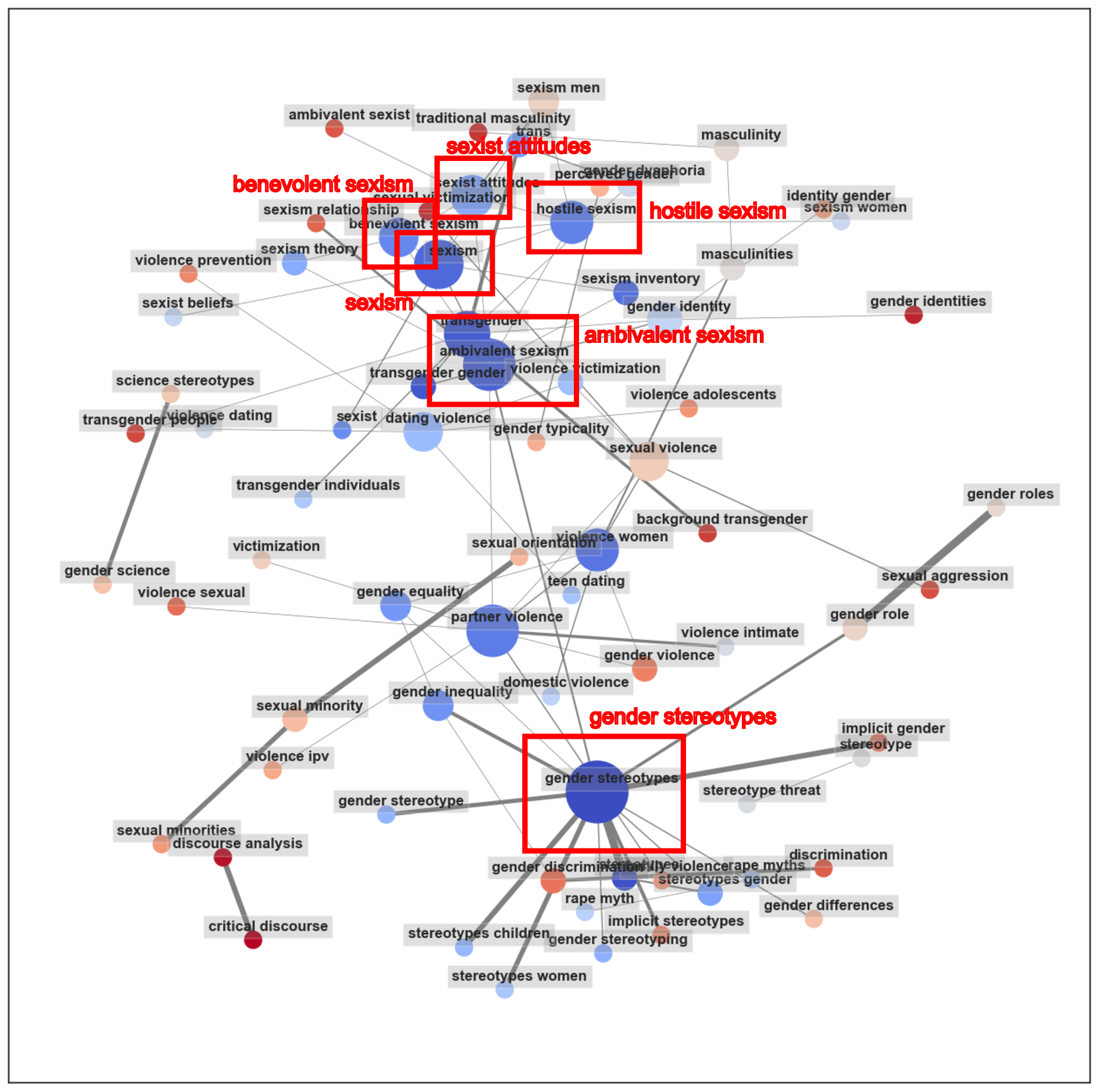}
    \caption{\centering Network diagram of most frequent keywords in \textbf{Computer Science}. \hspace{\textwidth} Among all the top 100 frequent and relevant keywords, the 6 most common ones (in descending order) are highlighted in the figure: 1. gender stereotypes 2. ambivalent sexism 3. hostile sexism \hspace{\textwidth}4. sexism 5. benevolent sexism 6. sexist attitudes}
    \label{fig:keywords_comp}
\end{figure}
In the Figure \ref{fig:keywords_comp}, we have the network diagram visualizing the most frequent and relevant keywords in Computer Science literature related to sexism and gender bias. The top 5 most common keywords are labelled in red boxes. Among the top 100 identified keywords, six emerged as particularly dominant: \textbf{gender stereotypes, ambivalent sexism, hostile sexism, sexism, benevolent sexism, and sexist attitudes}. These keywords are centrally positioned and form dense clusters, indicating their strong interconnections within the literature. The prominence of both psychological constructs (e.g., ambivalent sexism) and sociocultural phenomena (e.g., gender stereotypes) suggests a multidisciplinary engagement with sexism-related topics. The diagram highlights how research in Computer Science often intersects with gender theory, emphasizing the need for nuanced understandings of sexism in digital and computational contexts.

\hfill\break
\subsection{Most frequent keywords in Social Science}
\begin{figure}[htbp]
    \centering   
    \includegraphics[width=0.9\textwidth, keepaspectratio]{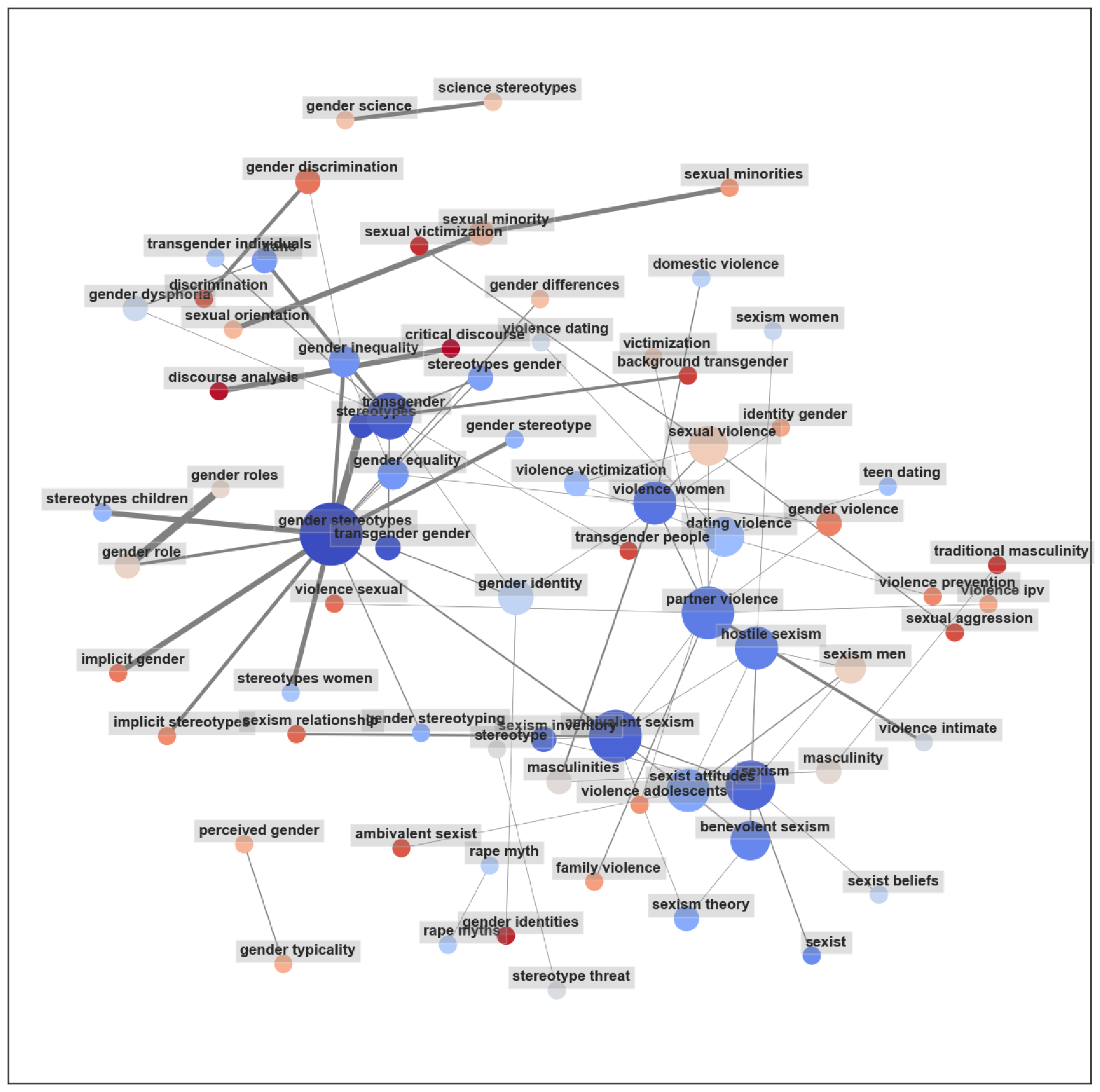}
    \caption{\centering Network connection of most frequent keywords \hspace{\textwidth}in Social Science}
    \label{fig:keywords_soc}
\end{figure}
In the Figure \ref{fig:keywords_soc}, the network diagram for Social Science presents the most frequent keyword connections related to sexism and gender bias within the discipline. Central nodes such as gender stereotypes, gender equality, gender identity, and gender roles indicate the strong thematic focus on societal structures and identity constructs. Compared to Computer Science, the Social Science network is more densely interlinked across themes like violence against women, transgender identity, and critical discourse, reflecting a broader and more intersectional engagement with gendered experiences. While terms like hostile sexism and benevolent sexism still appear, their relative positioning suggests they are less dominant than structural and sociological concepts. This visualization underscores Social Science’s emphasis on the sociocultural and identity-driven dimensions of sexism.

\section{Expanding on the automated selection techniques used}

\subsection{Topic Modeling approach}\label{sec:topic_mod_approach}

The model starts by transforming the input documents (abstracts and the titles) into numerical representations, with the help of embedding, which in these cases is a sentence embedding. Sentence embedding with transformer models maps a text of variable length to a fixed size embedding that should be representative of the the meaning of the input text. For our research, we used the sentence transformer `bge-small-en-v1.5'\footnote{The huggingface page of the model: \url{https://huggingface.co/BAAI/bge-small-en-v1.5}}, which maps the each paragraph of our document to a 384 dimensional dense vector space, that was then used to cluster topics of similar semantic structure. In topic modeling, it is key to have a good quality of topic representations to interpret the overall topic and understand patterns in the document, for which we used bag-of-words (BoW) of medium length n-gram value (1-3 n-grams). To further enhance the representative-ness of the topics from BoW, Term Frequency-Inverse Document Frequency (TF-IDF) of our document, which works on a document-level, were adjusted to c-TF-IDF as per their weights, which works on a cluster/categorical/topic level. It considers the differences in documents from different clusters, and can be calculated as: \\
c-TF-IDF (for a term \textcolor{blue}{x} within class \textcolor{blue}{c})
\begin{equation}
    W_{x, c} = ||tf_{x, c}|| \times log(1 + \frac{A}{f_{x}} )
\end{equation} where
\hspace{30pt}$tf_{x, c}$ = frequency of word \textcolor{blue}{x} in class \textcolor{blue}{c}, \\
\hspace*{2cm}$f_{x}$ = frequency of word \textcolor{blue}{x} across all classes, \\
\hspace*{2cm}A = average number of words per class \\
Though both of these approaches did a good job of acquiring the topic representations, we used representation models to fine-tune the topics to refine its representations. For that, we used a combination of three models - a fast keyword extraction model called KeyBERTInspired, PartOfSpeech model, and MaximalMarginalRelevance model. The KeyBERTInspired model increases the coherence and reduces stopwords 

Alongside this approach, we tried to further refine our topic representation by fine-tuning using a Large Language model (LLM) named `Mistral 7B v0.1' - a 7 billion parameter language model, which has shown to outperform other state-of-the-art language models like Llama 13B across all elevated benchmarks \cite{mistral}.

\begin{tcolorbox}
\begin{verbatim}
# The main representation of a topic
main_representation = KeyBERTInspired()

# Additional ways of representing a topic
pos_patterns = [
            [{'POS': 'ADJ'}, {'POS': 'NOUN'}],
            [{'POS': 'NOUN'}], [{'POS': 'ADJ'}]
]
aspect_model1 = PartOfSpeech("en_core_web_sm",
            pos_patterns=pos_patterns) 
aspect_model2 = [KeyBERTInspired(top_n_words=30, random_state=1234), 
            MaximalMarginalRelevance(diversity=.5)]

# LLM model
llm = Llama(model_path="../openhermes-2.5-mistral-7b.Q3_K_M.gguf", 
            n_gpu_layers=-1, n_ctx=4096, stop=["Q:", "\n"])
prompt = """ Q:
I have a topic that contains the following documents:
[DOCUMENTS]

The topic is described by the following keywords: '[KEYWORDS]'.

Based on the above information, can you give a short label
of the topic of at most 5 words?
A:
"""
aspect_model3 = LlamaCPP(llm, prompt=prompt)

# Add all models together to be run in a single `fit`
representation_model = {
           "Main": main_representation,
           "Aspect1":  aspect_model1,
           "Aspect2":  aspect_model2,
           "Aspect3": aspect_model3
        }
# The documents to train on are the titles and abstracts of the studies
topic_model = BERTopic(representation_model=representation_model)
                .fit(docs)
\end{verbatim}
\end{tcolorbox}

To assess the model performance\label{text:assessing_model}, the metrics perplexity and coherence scores were calculated as well. Perplexity is a predictive likelihood that specifically measures the probability that new data occurs given what was already learned by the model. In other words, perplexity characterizes how surprised a model is with new, unseen data. Coherence is typically used to analyze the relationship between two sets of data or the similarity between data sets. In topic modeling, topic coherence
measures the quality of the data by comparing the semantic similarity between highly repetitive words in a topic. We used this to maximize intra-topic and minimize inter-topic similarity. We attained a perplexity score of $1.23$ and a coherence score of $0.35$ from our topic model.

\section{Analysis of the Computer Science studies - the final selection}

In this section, we explore the data statistics for the CS manuscripts which were finally selected before the full-text screening process.
The visual analysis is purely based on the text contained in abstracts and titles of the selected studies.

\subsection{Documents by models}
\begin{figure}[h!]
    \centering   
    \includegraphics[width=0.9\textwidth, keepaspectratio]{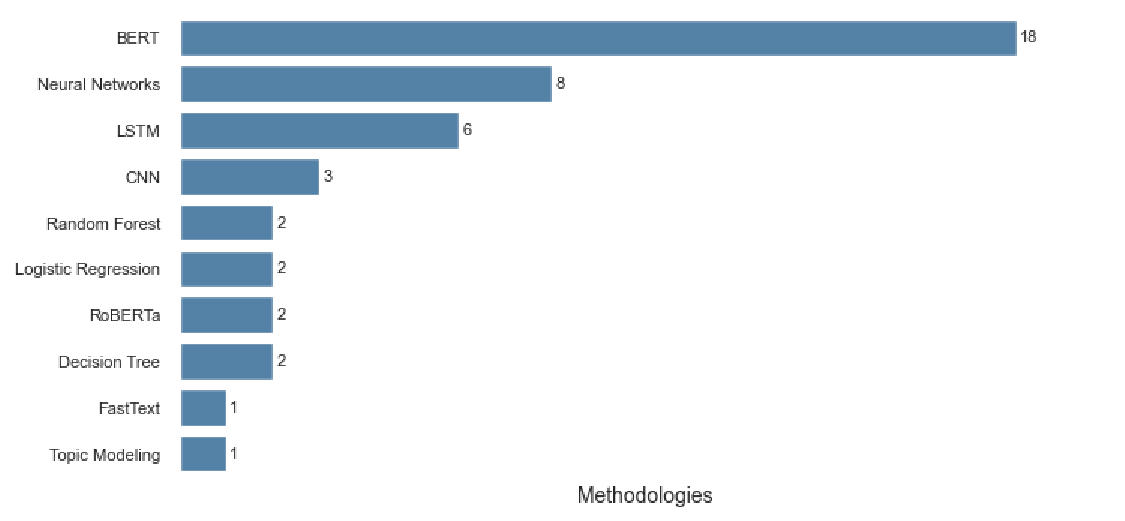}
    \caption{\centering Models gathered from the abstracts and titles of \hspace{\textwidth} Computer Science studies}
    \label{fig:models_comp}
\end{figure}

\hfill\break
\subsection{Task types and social platforms it is experimented on}\label{fig:task_type_platform}

\begin{figure}[h!]
  \centering
    \begin{minipage}[b]{0.4\textwidth}
        \includegraphics[width=\textwidth,height=6cm, keepaspectratio]{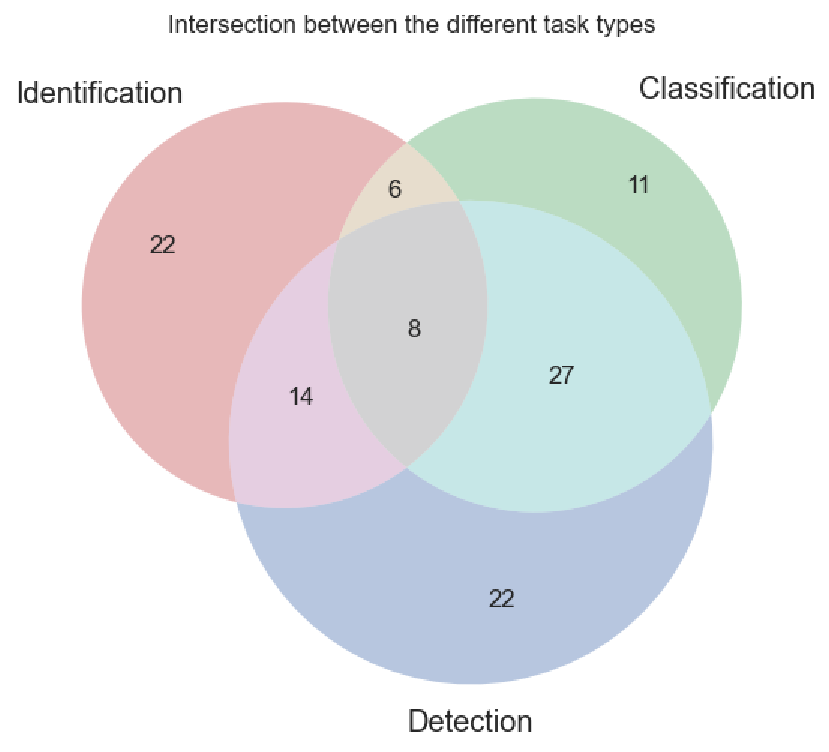}
        \caption*{\centering (a)}
        \label{img:task_type}
      \end{minipage}
      \hspace{-10pt}
      \hfill
      \begin{minipage}[b]{0.45\textwidth}
        \includegraphics[width=\textwidth,height=6cm]{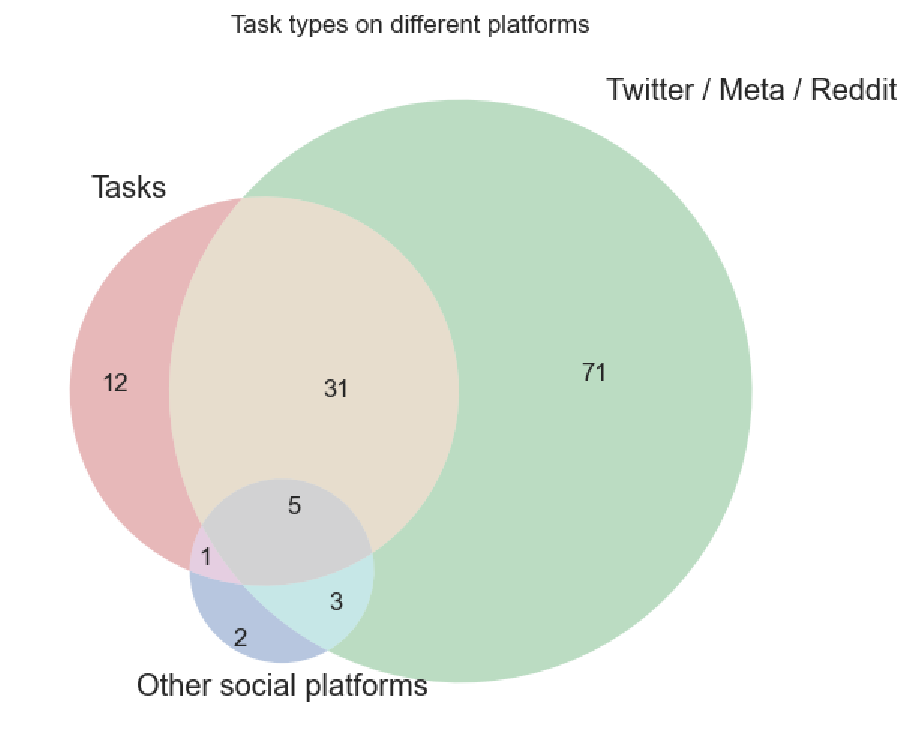}
        \caption*{\centering (b)}
        \label{img:task_platform}
    \end{minipage}
    \hfill
      \caption{\centering(a) Types of task present in CS for the quantification of sexism and misogyny. The task types in the figure represent the tasks that we ideally expect a paper to have when quantifying the said terms. Note that, ALL the three task types are relevant for our work, and there are 110 total. \hspace{\textwidth}(b) The aforementioned tasks and their application on different social media platforms. In this case, Twitter, Reddit and Facebook has shown to be the most research upon, while the other platforms are not. Regardless, a good number of research on those platforms use the specified tasks.}
\end{figure}

\section{Studies included in the Meta-analysis}
\begin{table}[ht!]
    \centering
    \begin{adjustbox}{width=\textwidth}
    \begin{tabular}{|p{4cm}|p{12cm}|}
    \hline
    \textbf{Citation} & \textbf{Title} \\
    \hline
    \citet{fersini_2018} & {Deep Learning Representations in Automatic Misogyny Identification: What Do We Gain and What Do We Miss?} \\
    \citet{rahali_akhloufi_2021} & {Automatic Misogyny Detection in Social Media Platforms using Attention-based Bidirectional-LSTM} \\
    \citet{frenda_2019} & {Online Hate Speech against Women: Automatic Identification of Misogyny and Sexism on Twitter} \\
    {\citet{bhattacharya_etal_2020}} & {Developing a Multilingual Annotated Corpus of Misogyny and Aggression} \\
    \citet{waseem-2016} & {Are You a Racist or Am I Seeing Things? Annotator Influence on Hate Speech Detection on Twitter} \\
    \citet{gamback-sikdar-2017-using} & {Using Convolutional Neural Networks to Classify Hate-Speech} \\
    \citet{waseem-hovy-2016} & {Hateful Symbols or Hateful People? Predictive Features for Hate Speech Detection on Twitter} \\
    \citet{sen-etal-2022-counterfactually} & {Counterfactually Augmented Data and Unintended Bias: The Case of Sexism and Hate Speech Detection} \\
    \citet{samory_sen_2021} & {“Call me sexist, but...” : Revisiting Sexism Detection Using Psychological Scales and Adversarial Samples} \\
    \citet{anzovino_2018} & {Automatic Identification and Classification of Misogynistic Language on Twitter} \\
    \citet{jha-mamidi-2017} & {When does a compliment become sexist? analysis and classification of ambivalent sexism using twitter data} \\
    \citet{butt_2021} & {Sexism Identification using BERT and Data Augmentation-EXIST2021} \\
    \citet{katsarou_2021} & {Sentiment Polarization in Online Social Networks: The Flow of Hate Speech} \\
    \citet{parikh_2019} & {Multi-label categorization of accounts of sexism using a neural framework} \\
    \citet{abburi_2021} & {Fine-Grained Multi-label Sexism Classification Using a Semi-Supervised Multi-level Neural Approach} \\
    \citet{melville_2019} & {Topic modelling of every-day sexism project entries} \\
    \citet{sharifirad_2018} & {Boosting Text Classification Performance on Sexist Tweets by Text Augmentation and Text Generation Using a Combination of Knowledge Graphs} \\
    \citet{nozza_volpetti_2019} & {Unintended bias in misogyny detection} \\
    \citet{chiril_etal_2020} & {An Annotated Corpus for Sexism Detection in French Tweets} \\
    \citet{chiril-etal-2021} & {“Be nice to your wife! The restaurants are closed”: Can Gender Stereotype Detection Improve Sexism Classification?} \\
    \citet{zeinert-etal-2021} & {Annotating online misogyny} \\
    \citet{guest-etal-2021} & {An expert annotated dataset for the detection of online misogyny} \\
    \citet{grosz_conde_2020} & {Automatic Detection of Sexist Statements Commonly Used at the Workplace} \\
    \citet{shah-etal-2020} & {Detecting hate speech against women} \\
    \citet{cans_2018} & {Misogyny Identification Through SVM} \\
    
    \hline
    \end{tabular}
    \end{adjustbox}
    \hfill
    \caption{Documents used for Meta-analysis (Part-1)}
    \label{tab:metaanalysis_docs_1}
\end{table}

\begin{table}[ht!]
    \centering
    \begin{adjustbox}{width=\textwidth}
    \begin{tabular}{|p{4cm}|p{12cm}|}
    \hline
    \textbf{Citation} & \textbf{Title} \\
    \hline
    \citet{schütz_2022} & {Automatic sexism detection with multilingual transformer models} \\
    \citet{lynn_endo_2019} & {A Comparison of Machine Learning Approaches for Detecting Misogynistic Speech in Urban Dictionary} \\
    \citet{talavera_2021} & {System Description for EXIST Shared Task at IberLEF 2021: Automatic Misogyny Identification Using Pretrained Transformers} \\
    \citet{plaza_2021} & {Sexism Identification in Social Networks using a Multi-Task Learning System} \\
    \citet{singh-etal-2021} & {“Hold on honey, men at work”: A semi-supervised approach to detecting sexism in sitcoms} \\
    \citet{mishra_2019} & {Abusive language detection with graph convolutional network} \\
    \citet{bashar_2020} & {Misogynistic Tweet Detection: Modelling CNN with Small Datasets, in: Communications in Computer and Information Science} \\
    \citet{frenda_2018} & {Exploration of Misogyny in Spanish and English Tweets} \\
    \citet{gordeev-lykova-2020} & {BERT of all trades, master of some} \\
    \citet{frenda_2018} & {Exploration of Misogyny in Spanish and English Tweets} \\
    
    \citet{plaza_del_arco_2020} & {Detecting misogyny and xenophobia in Spanish tweets using language technologies} \\
    \citet{attanasio_2022} & {Entropy-based attention regularization frees unintended bias mitigation from lists} \\
    \citet{rodriguez_2020} & {Automatic Classification of Sexism in Social Networks: An Empirical Study on Twitter Data} \\
    \citet{attanasio_2020} & {PoliTeam @ AMI: Improving Sentence Embedding Similarity with Misogyny Lexicons for Automatic Misogyny Identification in Italian Tweets} \\
    \citet{indurthi_etal_2019} & {FERMI at SemEval-2019 Task 5: Using Sentence embeddings to Identify Hate Speech Against Immigrants and Women in Twitter} \\
    \citet{kohli_etal_2021} & {ARGUABLY at ComMA@ ICON: Detection of Multilingual Aggressive, Gender Biased, and Communally Charged Tweets using Ensemble and Fine-Tuned IndicBERT} \\
    \citet{ou_li_2020} & {YNU\_OXZ@ HaSpeeDe 2 and AMI: XLM-RoBERTa with ordered neurons LSTM for classification task at EVALITA 2020} \\
    \citet{dinan_fan_et_al_2020} & {Multi-dimensional gender bias classification} \\
    \citet{francimaria_2022} & {Unintended Bias Evaluation: An Analysis of Hate Speech Detection and Gender Bias Mitigation on Social Media Using Ensemble Learning} \\
    \citet{Badjatiya_2017} & {Deep Learning for Hate Speech Detection in Tweets} \\
    \citet{fersini2021deep} & {Deep Learning Representations in Automatic Misogyny Identification: What Do We Gain and What Do We Miss?} \\
    \citet{farrell_fernandez_2019} & {Exploring Misogyny across the Manosphere in Reddit} \\
    \hline
    \end{tabular}
    \end{adjustbox}
    
    \caption{Documents used for Meta-analysis (Part-2)}
    \label{tab:metaanalysis_docs_2}
\end{table}




\end{document}